\PassOptionsToPackage{table}{xcolor}
\documentclass[]{fairmeta}

\usepackage{latexsym}
\usepackage{amsmath}
\usepackage{amssymb}
\usepackage{amsthm}
\usepackage{wrapfig}
\usepackage{enumitem}
\usepackage{url}
\usepackage{algorithm}
\usepackage{algpseudocode}
\usepackage{tabularx}
\usepackage{array}
\usepackage{float}
\usepackage{lineno}
\usepackage{titletoc}

\usepackage{amsmath,amsfonts,bm}

\def\eqref#1{equation~\ref{#1}}

\def\1{\bm{1}}

\DeclareMathAlphabet{\mathsfit}{\encodingdefault}{\sfdefault}{m}{sl}
\SetMathAlphabet{\mathsfit}{bold}{\encodingdefault}{\sfdefault}{bx}{n}

\tcbuselibrary{breakable,skins}

\microtypesetup{expansion=false}

\setlist{topsep=4pt, itemsep=2pt, parsep=0pt}

\definecolor{LT}{HTML}{EAF2F8} 
\definecolor{WT}{HTML}{FDEBD0} 
\definecolor{YC}{HTML}{E8F5E9} 

\definecolor{statebg}{HTML}{E3F1E0}   
\definecolor{histbg}{HTML}{FBEFD5}    
\definecolor{atombg}{HTML}{E1ECF7}    

\newcommand{\layerbox}[2]{{\setlength{\fboxsep}{1.2pt}\colorbox{#1}{#2}}}
\newcommand{\Slay}{\layerbox{statebg}{$\mathcal S_t$}}
\newcommand{\Hlay}{\layerbox{histbg}{$\mathcal H_t$}}
\newcommand{\Rlay}{\layerbox{atombg}{$\mathcal R_t$}}

\newcommand{\methodname}{\textit{MemRSI}}
\newcommand{\envname}{\methodname-Env}
\newcommand{\warmupname}{\methodname-Warmup}
\newcommand{\initname}{SC-GRPO}
\newcommand{\envshort}{Env}
\newcommand{\warmupshort}{Warmup}

\newcommand{\role}[1]{\textit{#1}}
\title{\methodname{}: Learning Self-Improving Memory Agents via Recursive Policy--Harness Evolution}
\renewcommand{\titlelist}{{\LARGE\sffamily\bfseries\setstretch{1.0}%
  \methodname{}: Learning Self-Improving Memory Agents\\via Recursive Policy--Harness Evolution\par}}

\author[2]{Litong Liu\textsuperscript{*}}
\author[1]{Tianxin Wei\textsuperscript{*\ensuremath{\dagger}}}
\author[3]{Yuanchen Bei}
\author[2]{Sikai Cheng}
\author[3]{Xuying Ning}
\author[3]{Zihao Li}
\author[3]{Zhining Liu}
\author[4]{Jiaru Zou}
\author[5]{Haonan Wang}
\author[3]{Zhenrui Yue}
\author[1]{Dongqi Fu}
\author[1]{Jianhui Sun}
\author[1]{Haixu Ma}
\author[1]{Yao Zhang}
\author[1]{Qifei Wang}
\author[1]{Ruiyi Li}
\author[1]{Serena Li}
\author[1]{Xiangjun Fan}
\author[1]{Bo Peng}
\author[1]{Weiwei Li}
\author[1]{Lizhu Zhang}

\affiliation[1]{Meta AI}
\affiliation[2]{Georgia Institute of Technology}
\affiliation[3]{University of Illinois Urbana-Champaign}
\affiliation[4]{Stanford University}
\affiliation[5]{National University of Singapore}

\contribution[*]{Equal contribution}
\contribution[\dagger]{Project lead.}

\date{November 26, 2026}
\correspondence{Litong Liu (\email{lliu607@gatech.edu}), Tianxin Wei (\email{tianxinwei@meta.com})}

\abstract{
Effective memory management enables personalized agents to use historical interactions and adapt to evolving user needs, and recent advances in reinforcement learning (RL) have opened new opportunities for learning memory management policies. However, memory policy learning typically relies on a harness designed in advance. As the policy improves, its remaining failure modes and learning needs change, while a fixed harness continues to organize interaction, feedback, and optimization according to its original design, which can limit further improvement.
To address this challenge, this paper introduces \methodname{}, a framework for learning self-improving personal memory agents through recursive policy--harness evolution.
\methodname{} builds a multimodal \textbf{Memory Environment} with a structured \emph{Memory Bank}, validated operations, source evidence, and outcome feedback; initializes the \role{manager} through \textbf{Memory Policy Learning} with SFT and \emph{State-Conditioned GRPO (SC-GRPO)}; and applies \textbf{Recursive Policy--Harness Evolution}, in which the \role{manager} diagnoses its own failures, revises the harness, and keeps a revision only when further learning improves memory utility.
On three main multimodal long-term conversational memory benchmarks, \methodname{} raises the average judge score by 8.6 points over the strongest controlled baseline.
A policy trained only on Mem-Gallery also improves over its untrained \role{manager} in all seven unseen evaluation settings without adaptation, by up to 11.3 points.
The benefits hold across 4B, 9B, and 27B \role{memory managers} and extend to diverse answer models, including proprietary LLMs such as GPT and Claude.

}

\begin{document}
\maketitle

\begin{figure}[!b]
  \centering
  \includegraphics[width=0.92\textwidth]{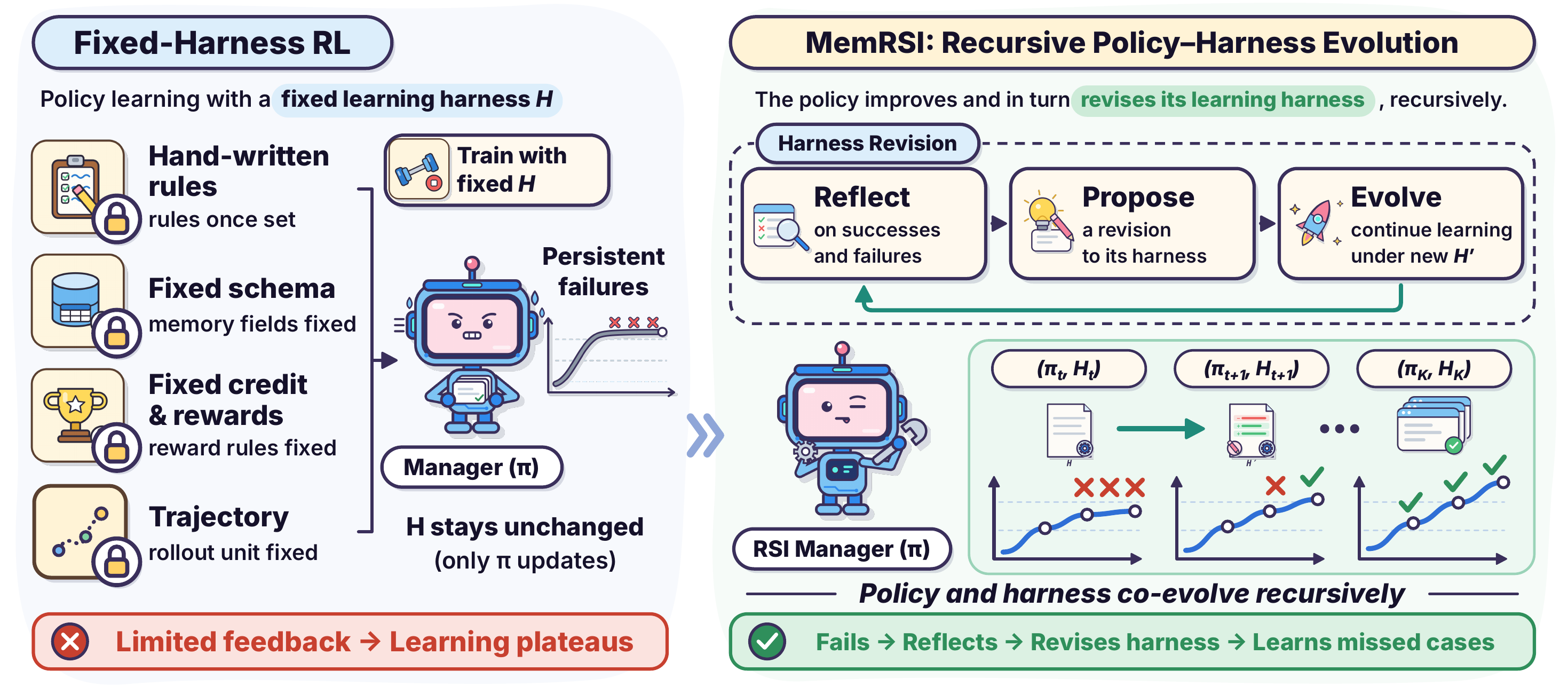}
  \caption{\textbf{Fixed-harness RL versus \methodname{}.}
    Left: the harness stays fixed while the policy improves.
    Right: \methodname{} uses the \role{manager}'s successes and failures to revise the harness, continues learning, and keeps the revised policy--harness pair when validation improves.}
  \label{fig:rl_rsi_motivation}
\end{figure}

\section{Introduction}
\label{sec:introduction}

Large language model (LLM) agents \citep{openai2023gpt4,yang2025qwen3} are increasingly deployed in long-horizon settings, such as software engineering \citep{yang2024sweagent}, web navigation \citep{zhou2024webarena}, and personal assistance \citep{wang2025mirix}.
These settings involve interactions that often extend across many sessions.
The need for memory is particularly pronounced in personal assistance, where information shared in early sessions may be needed only long afterward \citep{bei2026memgallery,maharana2024locomo}.
For example, a user may share a photograph of a plant, later correct its name, and describe a change in its condition weeks afterward.
In such settings, relying on the entire conversation history is insufficient,
as models become less reliable over long inputs
\citep{liu2024lost,wu2025longmemeval,li2025survey} and the raw history leaves the current state implicit.

Agents thus need a memory that selects and stores information across sessions. Earlier systems decide what to remember using hand-designed rules \citep{packer2023memgpt,chhikara2025mem0,wang2025mirix}. Following
the success of RL in post-training language models
\citep{ouyang2022training,guo2025deepseekr1}, recent methods train a memory
\role{manager} to make these decisions, using the accuracy of later answers as reward
\citep{yan2025memoryr1,wang2025memalpha}. Subsequent work refines this sparse
outcome reward with intermediate and decision-level signals
\citep{shen2026membuilder,ma2026finemem,yan2026memoryr2}.

However, continued memory-policy improvement faces a central challenge: the learning and testing loop is usually organized by a harness designed in advance, while the policy's needs evolve.
The harness determines how the \role{manager} interacts with memory, receives feedback, and turns experience into policy updates \citep{shao2024}.
As the policy improves, its remaining failure modes and learning needs change, but a fixed harness cannot adjust this guidance accordingly.
For memory agents, persistent state and interdependent operations make these needs especially dependent on current behavior: a failed answer may reflect an earlier write, a later update, retrieval, or answering \citep{liu2026worldmemarena}.
Connecting outcomes to memory interactions and supporting evidence makes it possible to identify where the harness should adapt \citep{sutton2018reinforcement}.

Memory research improves agents through several complementary routes.
Mem0 and A-Mem use designed prompts to extract, link, and revise memory entries \citep{chhikara2025mem0,xu2025amem}, while MemGPT, MIRIX, and M$^2$A organize memory interaction through hierarchies or specialized components \citep{packer2023memgpt,wang2025mirix,feng2026m2a}.
Memory-R1 and Mem-$\alpha$ train memory decisions from downstream outcomes \citep{yan2025memoryr1,wang2025memalpha}; MemBuilder, Fine-Mem, and Memory-R2 further improve the feedback used for learning these decisions \citep{shen2026membuilder,ma2026finemem,yan2026memoryr2}.
These methods improve memory content or policy behavior within a chosen prompt, workflow, or training design, while the procedure for further improving the \role{manager} is generally prescribed (Fig.~\ref{fig:rl_rsi_motivation}, left).
What remains less explored is how the \role{manager}'s own memory experience can guide revisions to that design as its limitations change.
Recursive self-improvement (RSI) offers a general route to this adaptation \citep{good1966speculations,schmidhuber2007godel}: the current \role{manager} identifies emerging weaknesses from memory construction and use, proposes a harness revision, and learns from the revised process.

To address this mismatch, this paper introduces \methodname{}, a unified framework for learning self-improving personal memory agents through recursive policy--harness evolution (Fig.~\ref{fig:rl_rsi_motivation}, right).
Following Memory-R1~\citep{yan2025memoryr1}, \methodname{} assigns memory construction and question answering to two models.
A trainable \role{memory manager} writes a persistent Memory Bank, and a frozen \role{answer agent} reads the bank to answer questions.
The Memory Environment integrates multimodal observations, Memory Bank, validated state transitions, source-evidence access, and outcome feedback.
The Memory Bank organizes current states, historical episodes, and atomic evidence; execution and feedback records make the \textit{manager}'s behavior available for diagnosis.
Within this environment, Memory Policy Learning initializes the \textit{manager} through SFT and RL.
Recursive Policy--Harness Evolution then adapts the harness to the policy's current learning needs: the \textit{manager} diagnoses its experience, proposes a revision, and learns under the revised harness (Fig.~\ref{fig:overview}).
Each revision is judged by the memory utility obtained after further learning under a fixed read protocol.
Successful policy--harness pairs guide subsequent rounds, keeping the improvement mechanism responsive to the policy it trains.

Extensive experiments on eight evaluation settings against eleven memory baselines demonstrate the scalability of \methodname{} across 4B, 9B, and 27B \role{memory managers}, with an average gain of 8.6 judge points over the strongest baseline on the three main multimodal long-term conversational memory benchmarks.
Recursive evolution further improves the outcome-trained 9B and 27B \role{managers}. The evolved 9B policy generalizes to seven additional evaluation settings without adaptation.
The resulting memories also benefit diverse answer models, including proprietary LLMs such as GPT and Claude.

Main contributions are summarized as follows:
\begin{itemize}[itemsep=0pt,topsep=2pt,leftmargin=*]
\item \textbf{Stateful Multimodal Memory Environment.} We integrate a structured Memory Bank, validated memory interactions, original-evidence access, and controlled outcome feedback, providing a foundation for policy learning and diagnosis of the agent's changing needs.
\item \textbf{Recursive Policy--Harness Evolution.} We introduce agent-driven revisions grounded in the current \role{manager}'s memory experience, retaining the revised policy--harness pair only when further learning improves memory utility and inheriting useful changes into subsequent rounds.
\item \textbf{Extensive Empirical Validation.} Experiments demonstrate consistent improvements across \role{memory managers} from 4B to 27B parameters, strong transfer to seven additional evaluation settings without adaptation, and generalization across diverse answer models, including GPT and Claude.
\end{itemize}

\section{Related Work}
\label{sec:related_work}

\paragraph{{Memory for multimodal agents.}}
Agent memory has progressed from preserving long histories to organizing and learning reusable knowledge \citep{huang2026survey}.
Hierarchical and summarization-based systems extend usable context \citep{packer2023memgpt,zhong2024memorybank,park2023generative}, while Mem0, A-Mem, and Zep maintain editable, linked entries \citep{chhikara2025mem0,xu2025amem,rasmussen2025zep}.
Multimodal systems organize textual and visual knowledge and recover original evidence when summaries are insufficient \citep{wang2025mirix,feng2026m2a,zhu2026tiermem}; other methods learn from audiovisual streams or anticipate the evidence needed for future questions \citep{long2025m3agent,zou2026taskmem,sun2026pmmc}.
Memory benchmarks expose the demands of temporal reasoning, knowledge updates, visual grounding, and evidence recovery \citep{maharana2024locomo,wu2025longmemeval,bei2026memgallery,guo2026memeye,liu2026worldmemarena}.
Following advances in RL post-training \citep{ouyang2022training,shao2024,guo2025deepseekr1}, learned \role{managers} optimize memory decisions from downstream feedback \citep{yan2025memoryr1,wang2025memalpha,yu2025memagent, du2026memoryt1}.
Long-horizon optimization and memory-specific methods improve feedback and credit assignment \citep{zhou2024archer,feng2025gigpo,shen2026membuilder,ma2026finemem,yan2026memoryr2}, while CMI-Mem supplements QA rewards with a query-independent information signal \citep{wang2026cmimem}.
Memory construction and reading can themselves adapt: MemSkill evolves the skills that guide extraction, and EvolveMem revises its retrieval configuration from diagnosed failures \citep{zhang2026memskill,liu2026evolvemem}.
These advances largely focus on memory organization, retrieval, or valuation under a harness designed in advance.
Yet memory operations induce persistent, interdependent state changes: a later update can invalidate an earlier write, and a visual summary can omit evidence needed by a future question.
Stored content and aggregate answer scores alone cannot reveal which operation introduced a failure or which source observation it misrepresented.
\methodname{} links current state, historical changes, and atomic multimodal evidence to the operations that produced them, making downstream failures traceable to memory decisions and their original sources.

\paragraph{{Self-improving agents.}}
Self-improving agents have progressed from accumulating experience to revising their own design and updating their own weights.
With model weights fixed, agents accumulate reflections, insights, and reusable strategies \citep{shinn2023reflexion,zhao2024expel,suzgun2026dc,zhang2025ace,ouyang2026reasoningbank}, and this stored experience supports continual learning \citep{zheng2026lifelong,wei2025evomemory}; MemEvolve further evolves the memory design that holds it \citep{zhang2026memevolve}.
Beyond stored experience, agents search over prompts, designs, or executable agent code \citep{yang2024opro,fernando2023promptbreeder,hu2025adas,zhang2025dgm,yin2025godel,zelikman2024stop}.
Recent work extends this search to agent harnesses \citep{lee2026recursiveharness,lee2026metaharness,zhang2026selfharness,wei2026evo,ning2026evoharness}, including over continual task streams \citep{xiao2026reharness,liu2026adaptiveautoharness}; SIM \citep{yu2026sim} records the rationale and outcomes of earlier harness edits so that optimization can continue across benchmarks and policy models.
Self-improvement also reaches the model weights: SkillRL evolves skills alongside policy training \citep{xia2026skillrl}, and Self-Improving Memory \citep{chen2026selfimprovingmemory} has a model rewrite its own memory instructions and then distills the improved behavior into its weights.
Terminal-RSI makes this process recursive, writing experience back into the model so that each improved terminal agent helps train its successor \citep{yang2026terminalrsi}.
The training procedure and the policy can also be adapted together, an idea with a long history in reinforcement learning: population based training evolves hyperparameters with policy weights \citep{jaderberg2017pbt}, AutoRL automates such design choices more broadly \citep{parkerholder2022autorl}, and Eureka lets an LLM write reward code and refine it from training feedback \citep{ma2024eureka}.
For LLM agents, harness adaptation is likewise coupled with policy training, as in Self-HPO, HarnessRL, and Continual Harness \citep{wei2026selfhpo,wang2026harnessrl,karten2026continualharness}.
Self-HPO selects executable revisions for the current policy before further training; HarnessRL adapts sampling, guidance, and rubric criteria; Continual Harness refines harness state online and updates the model from teacher-relabeled rollouts.
These frameworks use failure history to adapt execution scaffolds, experience stores, or task-level learning signals, but they are not designed for memory.
For personal memory, stored knowledge is shaped by interdependent operations, so revising the harness well requires evidence that traces outcomes back to those operations.
\methodname{} defines a memory-specific recursive framework in which the current \role{manager} uses environment states, operation histories, and source evidence to revise its harness, which can cover how memory is built and read, what feedback is measured, and how that feedback updates the policy; the revised policy--harness pair is retained only when further learning improves utility under a fixed read protocol.

\providecommand{\bank}{memory bank}
\providecommand{\Bank}{Memory bank}
\providecommand{\BankTitle}{Memory Bank}
\providecommand{\optterm}[1]{#1}

\FloatBarrier
\section{Methodology}
\label{sec:methodology}

A memory policy trained under a fixed harness eventually stops improving: the feedback that helped it early on no longer matches the mistakes it still makes.
\methodname{} lets the policy revise that harness itself.
The Memory Environment records how each memory decision changes the bank and affects later answers (\S\ref{sec:memory-environment}); Memory Policy Learning trains a capable \role{manager} under the initial harness (\S\ref{sec:policy-training}); Recursive Policy--Harness Evolution then has the \role{manager} diagnose its remaining failures, propose a harness revision, and keep it only if learning under it improves memory quality (\S\ref{sec:harness-evolution}).

Throughout, $H$ denotes the agent harness, which governs how the \role{manager} interacts with memory, what feedback it receives, and how that feedback updates the policy; $\theta$ denotes the policy parameters.
Memory Policy Learning produces the initial pair $(\theta_0,H_0)$, and Recursive Policy--Harness Evolution revises $\theta$ and $H$ together.

\subsection{Stateful Multimodal Memory Environment}
\label{sec:problem-formulation}
\label{sec:memory-environment}

\begin{figure}[!t]
  \centering
  \includegraphics[width=\textwidth]{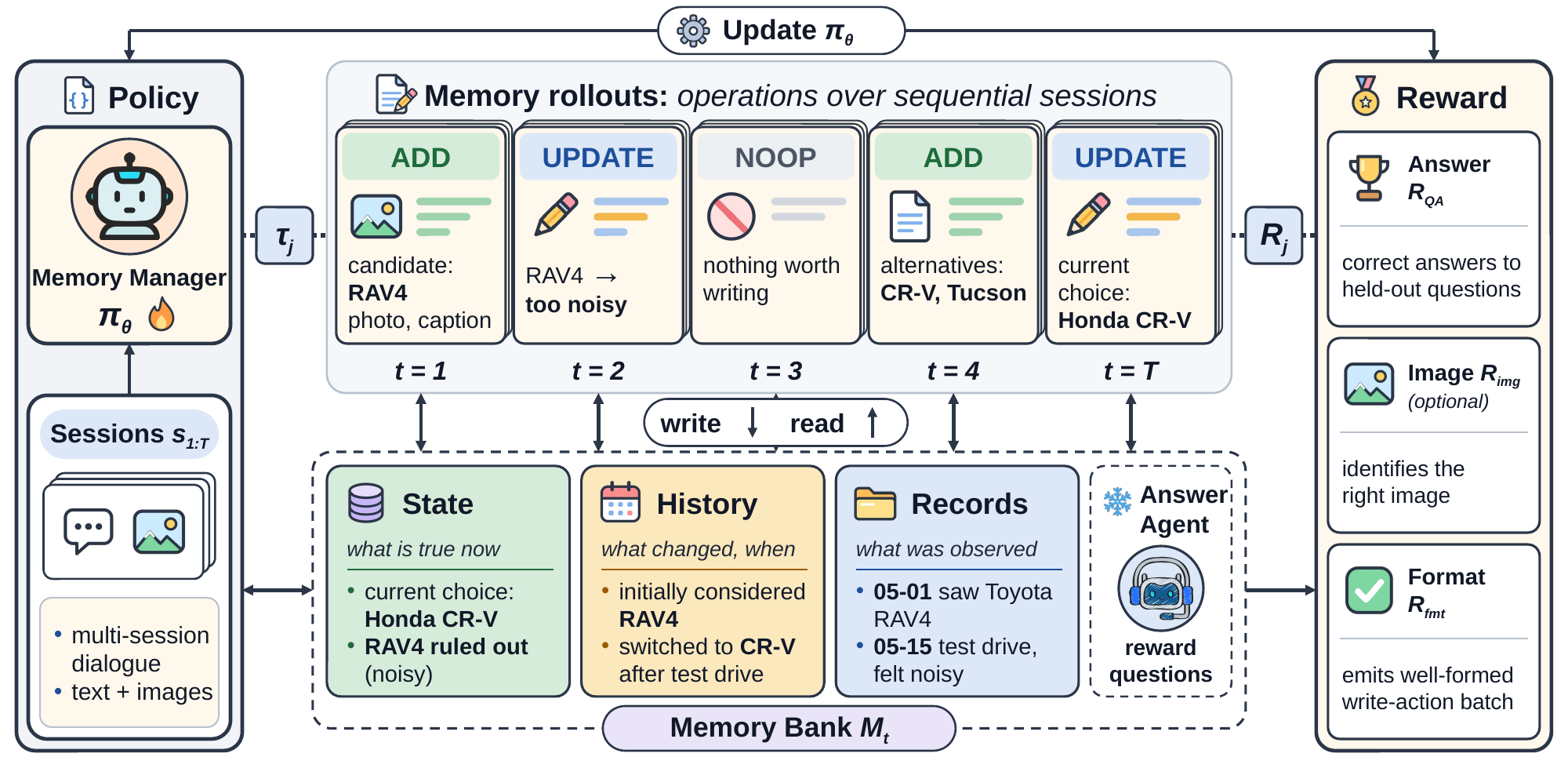}
  \caption{\textbf{Memory Environment.} The 
    \role{manager} writes one transaction per session to the Memory Bank, forming a write trajectory $\tau_j$. SC-GRPO compares transactions sampled for the same session from a shared memory state (\S\ref{sec:policy-training}), each scored by $R_{\mathrm{QA}}$, the optional $R_{\mathrm{img}}$, and $R_{\mathrm{fmt}}$.}
  \label{fig:memory_bank}
\end{figure}

In the Memory Environment, the \role{manager} reads each session and writes changes into a persistent Memory Bank; later, a fixed reader answers questions from that bank, and answer quality is the feedback (Fig.~\ref{fig:memory_bank}).
A deterministic \role{executor} applies validated changes, and every entry keeps links to its source evidence and to the operations that produced it, so outcomes can be traced back to decisions.

\paragraph{Memory Bank.}
\label{sec:memory-structure}

A personal memory must retain current knowledge without losing the historical and observational evidence that later questions may require.
A raw transcript keeps every observation but leaves the current state implicit \citep{maharana2024locomo}; a rolling summary keeps the current state but loses what it replaced \citep{packer2023memgpt}.

The \bank{} therefore separates what is true now, how it came to be, and what was actually observed (Fig.~\ref{fig:memory_bank}):
\begin{equation}
\mathcal M_t=(\Slay,\Hlay,\Rlay),
\label{eq:memory-layers}
\end{equation}
where (i) the \layerbox{statebg}{\emph{State}} \Slay{} holds the currently valid knowledge about each entity and keeps only the new value when a later observation supersedes an entry;
(ii) the \layerbox{histbg}{\emph{History}} \Hlay{} records what happened to the user as a chain of dated events across sessions, preserving an order that a single state entry cannot show;
and (iii) the \layerbox{atombg}{\emph{Records}} \Rlay{} store individual facts, instructions, and visual observations as atomic entries, each linked to the session it came from.

The \role{manager} described next maintains all three layers through \texttt{ADD}, \texttt{UPDATE}, \texttt{DELETE}, and \texttt{NOOP} operations~\citep{yan2025memoryr1}.

\paragraph{Memory management.}
In long-horizon personal assistance, a conversation is a sequence of sessions $s_{1:T}$, each containing dialogue and images.
The \emph{memory manager} $\pi_\theta$ (the \emph{manager} for short) reads each incoming session $s_t$ together with a view of the existing \bank{} $\mathcal M_{t-1}$.
It then writes a \emph{transaction} $a_t$, an ordered sequence of memory \emph{operations}. The \role{executor} validates operation types, layers, and source references, and applies accepted operations to the persistent bank:
\begin{equation}
\begin{aligned}
  a_t &\sim \pi_\theta(\cdot \mid s_t,\mathcal M_{t-1}),\\
  \mathcal M_t &= \operatorname{Execute}(\mathcal M_{t-1},a_t).
\end{aligned}
\label{eq:memory_policy}
\end{equation}
At question time, the environment's frozen \role{retriever} selects entries and their original source evidence from $\mathcal M_T$ for a question $q$, and an \role{answer agent} produces the answer from them.
The objective is to learn \role{manager} parameters $\theta$ that maximize the expected answer quality,
\begin{equation}
\max_\theta\; \mathbb{E}_{(s_{1:T},q)\sim p_{\mathrm{data}}}\,\mathbb{E}_{\tau\sim P_\theta(\cdot\mid s_{1:T})}\big[\,\operatorname{Score}(q,\mathcal M_T(\tau))\,\big],
\label{eq:objective}
\end{equation}
where $p_{\mathrm{data}}$ is the distribution of conversation--question pairs, $\tau=(a_1,\ldots,a_T)$ is the write trajectory generated by Eq.~\ref{eq:memory_policy}, and $\operatorname{Score}(q,\mathcal M_T(\tau))$ is the quality of the answer produced by the \role{answer agent} from $\mathcal M_T$.

\paragraph{Source-linked answering.}
Compressed entries may omit details a future question needs.
Each retrieved entry therefore carries source links that resolve the corresponding dialogue rounds and images, so the \role{answer agent} can inspect evidence that a summary may omit.

\paragraph{Provenance for self-improvement.}
A wrong answer may originate from an early omission, a later update, or a record that was never retrieved.
The final bank alone does not identify the responsible decision.
We therefore preserve source links, the operation history and replaced versions, and the entries exposed to the reader for each question.
With these records, a wrong answer can be traced back to the entries the reader saw, the operations that wrote them, and the observations they came from.
This reveals, for example, a description that lacks visual support, an update that overwrote earlier knowledge, or evidence that was never retrieved.
Because execution is deterministic, the trace also supports controlled tests: remove a suspected operation, replay the rest, and see whether the answer changes.

\begin{figure}[!t]
  \centering
  \includegraphics[width=\textwidth]{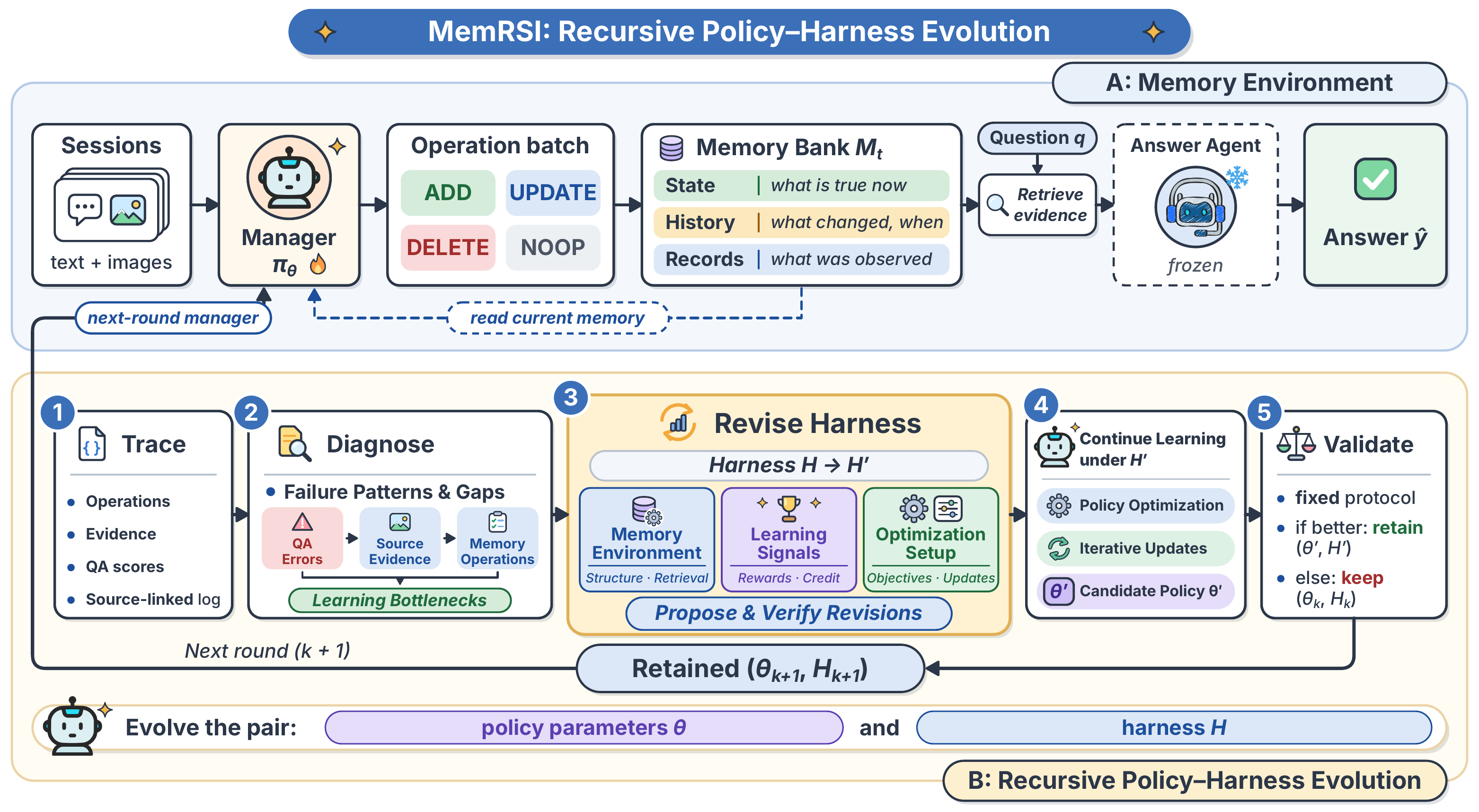}
  \caption{\textbf{Overview of \methodname{}.} (A) The
    \role{manager} builds source-linked memory for future answering. (B) Outcomes are traced to operations and evidence. The \role{manager} diagnoses them, proposes a harness revision, and keeps it if validation improves. Step 3 lists the revisable components: memory environment ($H^{\mathrm{mem}}$), learning signals ($H^{\mathrm{sig}}$), and optimization setup ($H^{\mathrm{opt}}$).}
  \label{fig:overview}
\end{figure}

\subsection{Memory Policy Learning}
\label{sec:policy-training}

Memory Policy Learning produces the initial pair $(\theta_0,H_0)$ in two steps.
SFT establishes valid transaction structure, layers, operations, and source references; SC-GRPO then trains preferences among executable memory decisions under the fixed initial harness $H_0$ until validation performance saturates.
We denote the resulting policy $E_0$; its parameters are $\theta_0$.
Its executed decisions and outcomes become the evidence from which the later revisions start.

\paragraph{Environment-generated SFT data.}
We generate SFT examples with Gemini 3.1 Pro inside the Memory Environment, so that demonstrations follow its observation and writing interface.
Each example pairs a multimodal session and its preceding memory state with a teacher-generated transaction in the environment's operation format.
The teacher uses the environment specification to generate memory updates with the appropriate layers and source references; audited transactions provide the supervision targets.
SFT optimizes transaction tokens while excluding the session and memory prompts from the loss.
This aligns the \role{manager} with executable memory behavior before downstream utility optimization.

\paragraph{State-Conditioned GRPO.}
A reliable comparison must isolate the current session's memory decisions from the history they inherit.
We instantiate Group Relative Policy Optimization (GRPO)~\citep{shao2024} at the session level, conditioning each rollout group on the same historical memory state. We call this \emph{State-Conditioned GRPO (SC-GRPO)}.
For session $s_t$, the historical environment state is built from preceding sessions and includes the current bank $\mathcal M_{t-1}$.
The \role{manager} generates $G$ independent rollouts from this shared history; each rollout consists of a transaction and its execution on a private copy of the environment state:
\begin{equation}
\begin{aligned}
a_t^{(j)}&\sim\pi_\theta(\cdot\mid s_t,\mathcal M_{t-1}),\\
\mathcal M_t^{(j)}&=\operatorname{Execute}(\mathcal M_{t-1},a_t^{(j)}),
\qquad j=1,\ldots,G.
\end{aligned}
\label{eq:session_rollouts}
\end{equation}
Thus candidates face the same observed past and differ in their current memory decisions.
Each candidate bank is then scored by how well the frozen \role{retriever} and \role{answer agent} answer the questions assigned to this session, $\mathcal Q_t$.
A question is assigned to the session in which its last supporting clue appears, and it is never shown to the \role{manager}.
Scoring $\mathcal M_t^{(j)}$ on $\mathcal Q_t$ is thus a session-level surrogate for the objective on the final bank $\mathcal M_T$ in Eq.~\ref{eq:objective}.
The reward for candidate $j$ is
\begin{equation}
R_j=R_{\mathrm{QA},j}
-R_{\mathrm{fmt},j}
\optterm{{}+\lambda_{\mathrm{img}} R_{\mathrm{img},j}},
\label{eq:memory-reward}
\end{equation}
where $R_{\mathrm{QA}}$ measures answer quality and $R_{\mathrm{fmt}}$ penalizes transactions that are not well formed.
\optterm{$R_{\mathrm{img}}$ is an optional term designed for tasks that depend heavily on visual evidence, weighted by $\lambda_{\mathrm{img}}$. It checks visual records against their source images and rewards descriptions that identify the right image among similar ones.}
Rewards are standardized within each group,
\begin{equation}
A_j=\frac{R_j-\bar R}{\operatorname{std}(R_{1:G})+\epsilon},
\qquad
\bar R=\frac{1}{G}\sum_{j=1}^{G}R_j,
\label{eq:group-advantage}
\end{equation}
and the advantage $A_j$ is applied to all tokens of transaction $j$ through the clipped GRPO objective with KL regularization.

Under $H_0$, what feedback is measured and how it is assigned are fixed before training begins.
As the policy improves, however, the mistakes it still makes change, and a feedback rule chosen in advance cannot follow them.
Once the policy saturates under $H_0$, further gains require revising that rule, and which revision helps can only be seen from the failures of the trained policy.

\FloatBarrier
\subsection{Recursive Policy--Harness Evolution}
\label{sec:harness-evolution}

Each round of Recursive Policy--Harness Evolution has four steps (Fig.~\ref{fig:overview}): the \role{manager} reads a report on its own failures, proposes a harness revision, trains under the revised harness for a fixed budget, and keeps the revision only if memory quality improves.
Because the proposer is the current \role{manager}, each revision targets the failures of the current policy.

\paragraph{Harness components.}
In existing memory-training procedures, the harness, including its feedback rules, is typically prescribed before optimization \citep{shen2026membuilder,ma2026finemem,yan2026memoryr2}.
We instead evolve the pair $(\theta,H)$, where the harness has three components,
$H=(H^{\mathrm{mem}},H^{\mathrm{sig}},H^{\mathrm{opt}})$, each a possible
revision target (Fig.~\ref{fig:overview}).
$H^{\mathrm{mem}}$ governs how the bank is constructed and how records are
retrieved (\S\ref{sec:memory-environment}).
$H^{\mathrm{sig}}$ determines what feedback is measured, such as which
outcomes are scored and how they become rewards.
$H^{\mathrm{opt}}$ determines how that feedback enters the policy update,
such as how rewards become advantages and how the update is regularized.
Which components are worth revising is left to the recursive process rather
than fixed in advance (\S\ref{sec:recursive_harness_adaptation}).
Learning under $H$ for $B$ optimizer updates is written
\begin{equation}
\theta'=\mathcal U_H^{(B)}(\theta).
\label{eq:harness-update}
\end{equation}
Across revisions, the update budget $B$ and the evaluation read path stay fixed.

\paragraph{Diagnosis from memory traces.}
The \role{manager} uses evidence from its current decisions to choose what to revise.
The report $\mathcal D_k$ combines training statistics with traces linking exposed entries to their writing operations, versions, and original observations.
These traces expose stale states, missing updates, or unsupported descriptions
and point to the harness component worth revising.

\paragraph{Revision hypotheses and executable constraints.}
The current \role{manager} turns its diagnosis into an executable revision:
\begin{equation}
r_k\sim\pi_{\theta_k}(\cdot\mid\mathcal{D}_k,H_k),
\qquad
H'_k=H_k\oplus r_k,
\label{eq:self-proposal}
\end{equation}
where $H_k\oplus r_k$ denotes the harness with the revision applied.
A proposal identifies its target weakness, changed component, and expected learning effect.
For example, a diagnosis that the bank confuses similar images could lead
to a revision of how records are constructed ($H^{\mathrm{mem}}$), of which
outcomes are scored ($H^{\mathrm{sig}}$), or of how the resulting rewards
update the policy ($H^{\mathrm{opt}}$); a proposal may change several
components at once.
Before policy learning, each revision is checked to run on the intended
decisions and produce finite learning signals; a revision that fails is
rejected without training.

\paragraph{Selection for learning gains.}
A revision's value shows only after the policy has learned under it, so selection compares memory utility after adaptation.
With the evaluation protocol held fixed, the gain of a candidate revision is
\begin{equation}
\Delta V_k(r_k)
=V\!\left(\mathcal U_{H_k\oplus r_k}^{(B)}(\theta_k)\right)-V(\theta_k).
\label{eq:adaptation_gain}
\end{equation}
In each round, one candidate that passes the checks is trained from the retained checkpoint.
Writing $\theta'_k=\mathcal U_{H'_k}^{(B)}(\theta_k)$, the \role{manager} and harness advance together only if the gain is positive:
\begin{equation}
(\theta_{k+1},H_{k+1})=
\begin{cases}
(\theta'_k,H'_k), & \Delta V_k(r_k)>0,\\[2pt]
(\theta_k,H_k), & \text{otherwise},
\end{cases}
\label{eq:frontier-update}
\end{equation}
where $V$ is a validation score on conversations held out from policy training and defined before evolution starts.
Reusing these conversations makes the gate a development signal; matched-continuation controls separately assess the effect of revision beyond the extra updates.
A rejected revision leaves the pair unchanged, and the next proposal starts from the same pair.
We index retained pairs as $E_0,E_1,\ldots$, so $E_i$ is the pair after the $i$-th accepted revision.
Algorithm~\ref{alg:coevolution} summarizes the loop.

\begin{algorithm}[t!]
\caption{Recursive memory-agent self-improvement}
\label{alg:coevolution}
\begin{algorithmic}[1]
\Require initial manager $\theta_0$ and harness $H_0$ (\S\ref{sec:policy-training}); update budget $B$; rounds $K$; validation score $V$
\For{$k = 0, \dots, K-1$}
  \State $\mathcal{D}_k \gets$ diagnostic report from the training rollouts of $\pi_{\theta_k}$, traced through the \bank{}
  \State $r_k \sim \pi_{\theta_k}(\cdot \mid \mathcal{D}_k, H_k)$ \Comment{manager proposes a revision}
  \State $H'_k \gets H_k \oplus r_k$ \Comment{implement}
  \If{$H'_k$ fails verification}
    \State $(\theta_{k+1}, H_{k+1}) \gets (\theta_k, H_k)$; \textbf{continue}
  \EndIf
  \State $\theta'_k \gets \mathcal{U}^{(B)}_{H'_k}(\theta_k)$ \Comment{train under the revised harness}
  \If{$V(\theta'_k) > V(\theta_k)$}
    \State $(\theta_{k+1}, H_{k+1}) \gets (\theta'_k, H'_k)$ \Comment{promote the pair}
  \Else
    \State $(\theta_{k+1}, H_{k+1}) \gets (\theta_k, H_k)$ \Comment{reject}
  \EndIf
\EndFor
\State \Return $(\theta_K, H_K)$
\end{algorithmic}
\end{algorithm}

\paragraph{Recursive coupling.}
Each improvement changes which failures remain, so the next proposal must come from the updated agent.
The retained \role{manager} stays the \role{proposer} in Eq.~\ref{eq:self-proposal}: its new memory outcomes refresh the diagnostic evidence, while $H_{k+1}$ carries the useful mechanisms forward.
Each revision therefore builds on both the current behavior and the revisions before it (\S\ref{sec:recursive_harness_adaptation}).

\section{Experiments}
\label{sec:experiments}

In this section, we conduct comprehensive experiments to address three research questions:
\begin{itemize}[leftmargin=*,itemsep=1pt,topsep=3pt]
  \item \textbf{RQ1: Memory quality.} Does \methodname{} outperform memory baselines, and how much does each of its components contribute?
  \item \textbf{RQ2: Generalization.} Does learned memory transfer across benchmarks, \role{answer agents}, \role{manager} sizes, and model families?
  \item \textbf{RQ3: Recursive evolution.} How much does Recursive Policy--Harness Evolution improve an SC-GRPO policy, and what changes drive those gains?
\end{itemize}

\subsection{Setup}
\label{sec:experimental_setup}

\paragraph{Benchmarks and protocol.}
Experiments are conducted on three main benchmarks: \textbf{Mem-Gallery}~\citep{bei2026memgallery} for long-term multimodal conversational memory, \textbf{MemEye}~\citep{guo2026memeye} for fine-grained visual evidence and temporal change, and \textbf{LoCoMo}~\citep{maharana2024locomo} for reasoning over extended multi-session dialogue.
Five further settings measure transfer without target training or checkpoint selection: \textbf{MemLens}~\citep{ren2026memlens}, \textbf{LongMemEval}~\citep{wu2025longmemeval}, \textbf{LongMemEval-v2 Web and Enterprise}~\citep{wu2026longmemevalv2}, and \textbf{PersonaMem}~\citep{jiang2025know}. Together with \textbf{MemEye} and \textbf{LoCoMo}, these give seven settings unseen by the Mem-Gallery policy.
The default \role{answer agent} is Qwen3.5-9B, with the same prompt and decoding for the controlled memory baselines; every answer is graded by a frozen Llama-3.3 judge.
\methodname{} retrieves ten evidence units with GME~\citep{zhang2024gme} and resolves their original source rounds.

\paragraph{Baselines and metrics.}
To evaluate the effectiveness of \methodname{}, we compare it with eleven representative memory baselines in two categories.
(i) \textit{Training-free memory systems}: \textbf{Full Memory (Text)}, \textbf{FIFO}, \textbf{RAG}~\citep{bei2026memgallery}, \textbf{Generative Agents}~\citep{park2023generative}, \textbf{MemGPT}~\citep{packer2023memgpt}, \textbf{Full Memory (Multimodal)}, \textbf{NGM}~\citep{fisher2025neural}, \textbf{AUGUSTUS}~\citep{jain2025augustus}, and \textbf{UniversalRAG}~\citep{yeo2026universalrag}. These methods cover full-history, structured memory, and retrieval-based approaches.
(ii) \textit{Trained memory managers}: \textbf{Memory-R2}~\citep{yan2026memoryr2} and \textbf{MemAgent}~\citep{yu2025memagent}, which optimize memory management through learned policies.
For fair comparisons, all eleven baselines use the same frozen \role{answer agent}, while retaining their respective memory construction and retrieval mechanisms.
We additionally include GPT-5.6 Luna and Claude Sonnet 5 as frontier model references, which perform memory writing and question answering using their own models and are reported separately from the controlled baselines.
The primary metric is the judge score, defined as the mean grade assigned by a frozen Llama-3.3 judge under a five-level rubric.
We report overall and question-family-level judge scores, while PersonaMem uses normalized multiple-choice exact match.

\paragraph{Implementation.}
The default \role{manager} is Qwen3.5-9B, and Gemma4-4B and Qwen3.8-27B test other sizes and model families.
We compare four \role{manager} stages, each adding one component: \envname{} (Memory Environment; Env in compact plots), \warmupname{} (+SFT), SC-GRPO ($E_0$), and \methodname{} (+Recursive Policy--Harness Evolution).
In the main table, \methodname{} denotes the final policy trained on each benchmark.
The transfer study uses only the Mem-Gallery-trained policy.
Full settings appear in Appendix~\ref{app:reproducibility_controls}.

\subsection{Memory Quality (RQ1)}
\label{sec:main_results}

\begin{table}[!t]
\centering\scriptsize
\setlength{\tabcolsep}{6.0pt}
\renewcommand{\arraystretch}{1.04}
\begin{tabular}{l r r r >{\columncolor{black!4}}r r r r >{\columncolor{black!4}}r r r r >{\columncolor{black!4}}r}
\toprule
\multirow{2}{*}{\textbf{Method / Model}}
& \multicolumn{4}{c}{\textbf{Mem-Gallery}}
& \multicolumn{4}{c}{\textbf{MemEye}}
& \multicolumn{4}{c}{\textbf{LoCoMo}} \\
\cmidrule(lr){2-5}\cmidrule(lr){6-9}\cmidrule(lr){10-13}
& Ext. & Reas. & Mgmt. & \textbf{Avg.}
& Y1 & Y2 & Y3 & \textbf{Avg.}
& Dir. & Comp.-T. & Open-R. & \textbf{Avg.} \\
\midrule
\multicolumn{13}{l}{\textbf{\textit{Frontier capacity references}}} \\
GPT-5.6 Luna & 89.4 & 74.1 & 86.9 & 85.2 & 64.8 & 62.6 & 74.8 & 64.6 & 76.5 & 70.4 & 59.6 & 70.0 \\
Claude Sonnet 5 & 86.2 & 84.8 & 81.7 & 85.2 & 62.4 & 52.8 & 64.5 & 57.2 & 76.5 & 71.7 & 70.0 & 73.2 \\
\midrule
\multicolumn{13}{l}{\textbf{\textit{Training-free memory methods}}} \\
Full Memory (Text) & 39.0 & 23.4 & 74.6 & 38.7 & 19.1 & 24.5 & 38.9 & 23.7 & 24.5 & 21.1 & 80.4 & 38.9 \\
FIFO & 20.5 & 3.7 & 70.9 & 24.7 & 28.4 & 23.2 & 33.9 & 26.1 & 3.4 & 5.1 & 84.5 & 25.6 \\
RAG & 50.7 & 57.0 & 82.1 & 56.3 & 32.5 & 29.4 & 52.7 & 32.9 & 58.1 & 44.4 & 79.5 & 60.2 \\
Generative Agents & 29.8 & 21.5 & 78.7 & 33.7 & 27.5 & 22.0 & 46.4 & 27.2 & 14.2 & 15.1 & 84.5 & 33.2 \\
MemGPT & 48.3 & 51.6 & 78.0 & 53.0 & 33.9 & 28.0 & 47.3 & 27.2 & 58.7 & 43.1 & 78.6 & 59.8 \\
Full Memory (Multimodal) & 34.6 & 11.2 & 67.2 & 33.5 & 33.9 & 29.8 & 45.5 & 32.4 & 13.8 & 11.4 & 80.6 & 32.3 \\
NGM & 88.1 & 70.1 & 87.3 & 83.3 & 74.8 & 52.4 & 79.8 & 62.6 & 48.7 & 35.6 & 79.8 & 53.3 \\
AUGUSTUS & 66.1 & 61.4 & 79.5 & 67.4 & 67.5 & 46.6 & 72.7 & 55.9 & 44.7 & 33.3 & 81.1 & 51.0 \\
UniversalRAG & 90.1 & 70.6 & 73.5 & 83.3 & 75.8 & 52.8 & 76.4 & 63.4 & 58.2 & 44.6 & 78.6 & 60.0 \\
\midrule
\multicolumn{13}{l}{\textbf{\textit{External trained managers}}} \\
Memory-R2 & 75.8 & 49.1 & 83.6 & 74.0 & 80.0 & 62.1 & 81.4 & 67.0 & 53.3 & 39.8 & 77.4 & 55.8 \\
MemAgent & 36.3 & 18.2 & 79.9 & 40.3 & 35.1 & 28.2 & 35.7 & 33.6 & 27.3 & 22.2 & 82.8 & 38.6 \\
\midrule
\multicolumn{13}{l}{\textcolor{green!45!black}{\textbf{\textit{Ours: Qwen3.5-9B memory policy}}}} \\
MemRSI-Env & 84.4 & 78.5 & 89.9 & 83.8 & 73.1 & 57.7 & 75.4 & 64.6 & 71.1 & 56.4 & 78.8 & 68.8 \\
MemRSI-Warmup & 85.8 & 77.8 & 82.1 & 83.2 & 73.4 & 63.8 & 75.0 & 68.1 & 72.9 & 60.9 & 80.2 & 71.3 \\
\rowcolor{YC}\textbf{MemRSI} & 89.3 & 85.5 & 87.3 & \textbf{88.0} & 80.4 & 64.0 & 75.5 & \textbf{70.6} & 76.3 & 64.5 & 80.7 & \textbf{74.0} \\
\bottomrule
\end{tabular}

\caption{\textbf{Answer quality on the three main benchmarks} (Llama-3.3 judge, \%). Ext./Reas./Mgmt.: extraction/reasoning/management; Y1--Y3: MemEye levels; Dir./Comp.-T./Open-R.: direct/compositional--temporal/open--robustness. Controlled methods share the Qwen3.5-9B \role{answer agent}; GPT and Claude write and answer with their own models. \envname{} uses the Memory Environment without post-training, and \warmupname{} adds SFT. Bold marks the best average, computed over all questions including categories not shown.}

\label{tab:main_results}
\end{table}

\textbf{The Memory Environment already gives a strong memory before any training.}
With no post-training, \envname{} matches or exceeds the strongest training-free method on all three benchmarks and outperforms both trained baselines on two of them (Table~\ref{tab:main_results}).
Its lead is largest on LoCoMo, where questions connect events scattered across many sessions and the dated History layer keeps them in order, and smallest on Mem-Gallery, where human-written captions already make visual evidence easy to find.

\textbf{\methodname{} achieves the best score on all three benchmarks.}
\methodname{} averages 77.5, 8.6 points above the strongest baseline, UniversalRAG, and also outperforms GPT and Claude, which write and answer with their own frontier models.
Where the gain comes from differs by benchmark.
On LoCoMo, the layered bank of \envname{} already supplies most of the margin; on Mem-Gallery, most of it comes from training, led by reasoning questions that combine several pieces of evidence (78.5 to 85.5).

\textbf{Sparse memories score well on questions that call for declining.}
Two question families reverse the usual ranking: Mem-Gallery management, which includes answer-refusal questions, and LoCoMo open--robustness, which includes questions the history cannot answer.
FIFO answers almost no direct LoCoMo question (3.4) yet has the best open--robustness score (84.5); on Mem-Gallery it scores 3.7 on reasoning but 70.9 on management.
A memory that keeps little gives the \role{answer agent} little to work with, so it often declines, and on these questions declining is correct.
A richer memory pulls the other way: when the bank holds many related observations, the \role{answer agent} more often finds something that looks relevant and answers instead of declining.
Accordingly, answer refusal is the category where \methodname{} drops most relative to \envname{} on Mem-Gallery (90.5 to 86.9; Table~\ref{tab:posttraining}), while refusals on questions that the evidence does support remain rare (Fig.~\ref{fig:failure_analysis}).
The lower scores on these families therefore reflect how much the bank exposes, not missing knowledge.

\begin{table}[!t]
\centering\small
\setlength{\tabcolsep}{4.5pt}
\renewcommand{\arraystretch}{1.15}
\begin{tabular}{lr@{\hspace{12pt}}lr@{\hspace{12pt}}lr}
\toprule
\multicolumn{2}{c}{\textbf{Memory Environment}} & \multicolumn{2}{c}{\textbf{Memory Policy Learning}} & \multicolumn{2}{c}{\textbf{Recursive Policy--Harness Evolution}} \\
\cmidrule(r){1-2}\cmidrule(lr){3-4}\cmidrule(l){5-6}
Free-form memory (w/o Environment) & 78.20 & + SFT & 83.22 & Matched continuation & 84.36 \\
Structured memory (\envname{}) & 83.80 & + Standard GRPO & 81.65 & Fixed best-evolved harness & 86.33 \\
& & + SC-GRPO ($E_0$) & 85.62 & \cellcolor{YC}Recursive evolution ($E_4$) & \cellcolor{YC}\textbf{88.03} \\
\bottomrule
\end{tabular}
\caption{\textbf{Component ablation on Mem-Gallery.} Judge scores (\%) with the Qwen3.5-9B \role{manager} and \role{answer agent}.}
\label{tab:component_ablation}
\end{table}

\textbf{Each of the three components of \methodname{} contributes, and each has a natural weaker alternative.}
Table~\ref{tab:component_ablation} replaces each component of \S\ref{sec:methodology} with a simpler choice on Mem-Gallery.
Letting the \role{manager} write free-form notes, without layers, validated operations, or source links, gives 78.2; the structured bank alone raises this to 83.8.
SFT then teaches the \role{manager} to write valid transactions, and SC-GRPO rewards banks that lead to correct answers~\citep{ranzato2016sequence}, reaching 85.6.
Standard GRPO, which rolls out whole conversations instead of single sessions from a shared state, produces long and noisy rollouts and stops at 81.7.
On top of the saturated policy, Recursive Policy--Harness Evolution adds 2.4 points, more than either fixed-harness control (\S\ref{sec:recursive_harness_adaptation}).

\textbf{The learned bank is both smaller and more useful than a retrieval store.}
UniversalRAG stores retrieval units for the whole history, whereas the \role{manager} decides what to keep, merge, or replace.
In twelve of fourteen histories, the learned bank uses fewer tokens, typically about half as many (median reduction 48.4\%), and it still answers better on every benchmark (Fig.~\ref{fig:memory_efficiency}).
On MemEye, where the two banks are about the same size, the learned bank still scores about 10\% higher, so the gain comes from what is stored rather than how much.

\subsection{Generalization (RQ2)}
\label{sec:generalization}

Beyond accuracy on the training benchmarks, a useful memory policy should also generalize to new histories, model backbones, and \role{answer agents}.
This section therefore applies the Mem-Gallery policy to unseen benchmarks, repeats policy learning with other \role{manager} backbones, and reads the learned memory with other \role{answer agents}.

\begin{table}[!t]
\centering\scriptsize
\setlength{\tabcolsep}{3.2pt}
\resizebox{\textwidth}{!}{%
\begin{tabular}{lcccccccc}
\toprule
\textbf{Method} & Mem-Gallery & MemEye & LoCoMo & MemLens & LongMemEval & LME-v2 Web & LME-v2 Enterprise & PersonaMem \\
\midrule
\rowcolor{YC} Ours (\methodname{}) & \textbf{88.0} & \textbf{69.7} & \textbf{71.9} & \textbf{63.1} & \textbf{81.6} & \textbf{53.9} & \textbf{52.7} & \textbf{72.7} \\
Ours (\envname{}) & 83.8 & 64.6 & 68.8 & 52.7 & 76.8 & 52.4 & 41.4 & 61.6 \\
\midrule
UniversalRAG & 83.3 & 63.4 & 60.0 & 59.4 & 72.4 & 26.8 & 27.8 & \textbf{72.7} \\
NGM & 83.3 & 62.6 & 53.3 & 57.4 & 63.8 & 22.0 & 19.3 & 72.5 \\
\bottomrule
\end{tabular}}
\caption{\textbf{Cross-dataset transfer under a fixed answer protocol.} One
Mem-Gallery-trained policy is reused without target adaptation. PersonaMem
reports normalized MCQ exact match; all other datasets report the judge score.}
\label{tab:cross_dataset_transfer_complete}
\end{table}

\textbf{One policy trained on Mem-Gallery improves memory in all eight settings.}
We apply the Mem-Gallery \role{manager} unchanged to each target history, with no target training or checkpoint selection.
It improves over \envname{} on Mem-Gallery and in all seven unseen settings and matches or exceeds UniversalRAG and NGM throughout (Table~\ref{tab:cross_dataset_transfer_complete}).
The gains are largest where histories are long or heterogeneous: 10.4 points on MemLens, 11.3 on LongMemEval-v2 Enterprise, and 11.1 points of exact match on PersonaMem (Fig.~\ref{fig:cross_dataset_generalization}).
LongMemEval-v2 shows the contrast most clearly: its web and enterprise histories are far from conversational, and the two retrieval baselines fall to 24.0 on average, consistent with the reported bias of retrieval pipelines toward conversational context~\citep{wu2026longmemevalv2}, while the learned memory stays at 53.3.
What transfers is therefore how the \role{manager} organizes a history, not a fit to particular questions.

\begin{figure}[!t]
  \newcommand{\alignedplot}[1]{%
    \parbox[c][0.70\linewidth][c]{\linewidth}{\centering
      \includegraphics[width=\linewidth,height=0.70\linewidth,keepaspectratio]{#1}}%
    \par}
  \begin{minipage}[t]{0.485\textwidth}
    \vspace{0pt}\centering
    \alignedplot{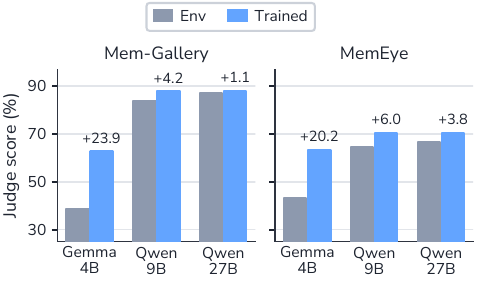}
    \caption{\textbf{Manager generalization.} Judge scores of \envshort{} and trained \role{managers}, each trained on the evaluated benchmark and read by the Qwen3.5-9B \role{answer agent}. Labels give gains over \envshort{}.}
    \label{fig:cross_backbone_final}
  \end{minipage}
  \hfill
  \begin{minipage}[t]{0.485\textwidth}
    \vspace{0pt}\centering
    \alignedplot{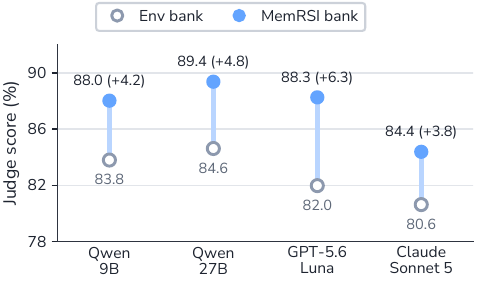}
    \caption{\textbf{Answer-agent generalization.} Mem-Gallery judge scores when four \role{answer agents} read the same evolved 9B bank (filled markers) or the matching \envshort{} bank (hollow markers). Labels give gains.}
    \label{fig:answer_agents}
  \end{minipage}
\end{figure}

\textbf{Policy learning benefits managers of different sizes and model families.}
With the \role{answer agent} fixed, training improves every \role{manager} on both Mem-Gallery and MemEye (Fig.~\ref{fig:cross_backbone_final}).
The smaller the \role{manager}, the larger the gain: Gemma4-4B improves by more than 20 points on both benchmarks, Qwen3.5-9B by 4--6, and Qwen3.8-27B by 1--4.
An untrained small model writes poor memory and has the most to gain, but even the 27B \role{manager}, which already writes a strong bank, still improves.

\textbf{The learned memory remains useful when the answer agent changes.}
We keep the evolved 9B bank fixed and change who reads it: two Qwen models, GPT, and Claude (Fig.~\ref{fig:answer_agents}).
Every answer agent scores 3.8 to 6.3 points higher with the learned bank than with the \envname{} bank.
The improvement therefore lies in the memory itself rather than in a fit to the 9B \role{answer agent} used during training.

\FloatBarrier
\subsection{Recursive Policy--Harness Evolution (RQ3)}
\label{sec:recursive_harness_adaptation}

The preceding results do not separate recursive evolution from the training that precedes it.
We therefore start from SC-GRPO policies trained to saturation ($E_0$) and ask two questions: does evolving the harness keep improving the policy, and is the gain due to the revisions rather than to further training?

\begin{figure}[!t]
  \centering
  \includegraphics[width=\linewidth]{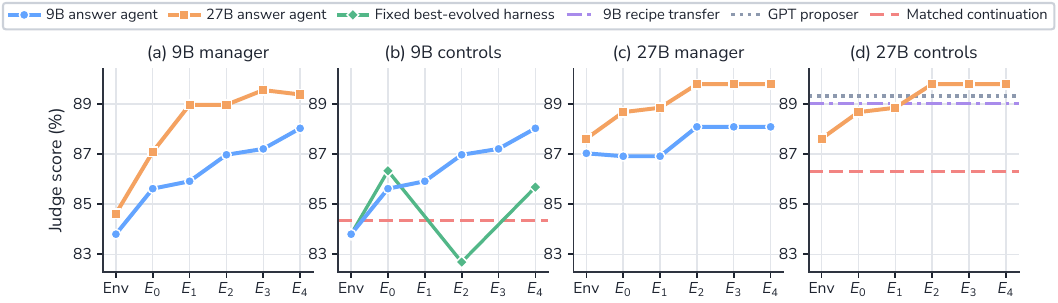}
  \caption{\textbf{Recursive Policy--Harness Evolution of 9B and 27B managers} (Mem-Gallery judge score, \%). \envshort{} is \envname{}, $E_0$ the SC-GRPO policy, and $E_1$--$E_4$ the retained revisions; the 27B \role{manager} keeps $E_2$ after the second round. (a,c) Scores under the 9B and 27B \role{answer agents}. (b) Retraining from SFT with the best-evolved harness fixed, plotted at the training lengths of $E_0$, $E_2$, and $E_4$. (d) Endpoints of the 9B recipe and GPT-proposed revisions after two rounds on 27B. Red dashed lines mark matched continuation.}
  \label{fig:learning_harness_evolution}
\end{figure}

\textbf{Recursive evolution keeps improving after fixed-harness training has saturated.}
For the 9B \role{manager}, each retained revision raises the score, from 85.6 at $E_0$ to 88.0 at $E_4$ (Fig.~\ref{fig:learning_harness_evolution}a).
This gain is larger than what SFT and SC-GRPO together add over \envname{}.
The 27B \role{manager} shows the same pattern more sharply: fixed-harness SC-GRPO leaves its score unchanged under the 9B \role{answer agent}, and the improvement comes only once the harness is revised (Fig.~\ref{fig:learning_harness_evolution}c).

\textbf{The gain comes from revising the harness, not from training longer.}
Two controls give the policy the same extra training without recursive revision (Fig.~\ref{fig:learning_harness_evolution}b,d).
Matched continuation keeps training $E_0$ under the initial harness, and the score drops below $E_0$ for both \role{managers}: further training under the initial harness brings no additional gain.
The second control takes the final evolved harness and trains a new policy with it from SFT.
If the evolved harness were simply a better fixed recipe, this should reach $E_4$.
Instead, training is unstable and never reaches $E_4$.
One likely reason is that each revision was designed for the failures of the policy at that point; applied from the start, it acts on a policy that has not yet developed the behavior those revisions were meant to refine.
The harness is useful because it evolves together with the policy.

\textbf{Revisions work best for the policy that proposed them.}
For the 27B \role{manager}, we compare its own revisions with two alternatives: applying the 9B revision sequence, and letting GPT propose revisions (Fig.~\ref{fig:learning_harness_evolution}d).
Both alternatives improve over $E_0$, but the 27B \role{manager}'s own revisions end highest.
Revisions found for the 9B \role{manager} only partly fit the 27B \role{manager}, whose failures differ.
GPT is a stronger general model, yet in this comparison its proposals end slightly below those of the 27B \role{manager} itself, suggesting that having the policy diagnose its own behavior and propose its own revisions can matter as much as the strength of the proposer.

\textbf{A stronger answer agent reveals more of the improvement.}
Revisions are selected with the 9B \role{answer agent}, yet the same banks also improve under the 27B \role{answer agent}, often by more (Fig.~\ref{fig:learning_harness_evolution}a,c).
For the 27B \role{manager}, the gain over \envname{} doubles when the 27B agent reads the bank, reaching 89.8.
The 9B agent thus cannot use all of what the improved memory contains, and selection with it, if anything, understates the improvement.

\begin{figure}[!t]
  \centering
  \includegraphics[width=\linewidth]{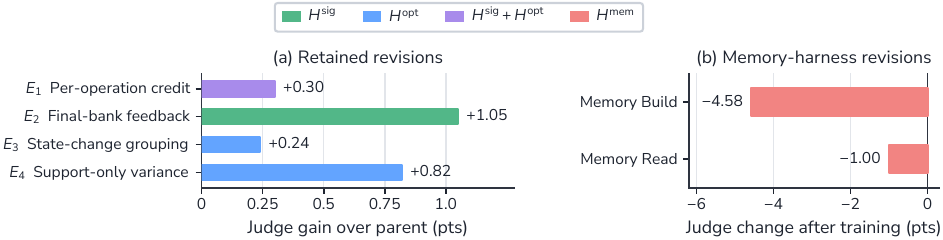}
  \caption{\textbf{Effects of recursive harness revisions.}
    (a) Judge gain of each retained revision over its parent, colored by the revised component. (b) Judge change after training under the two memory-harness ($H^{\mathrm{mem}}$) searches: Memory Build changes the bank structure, and Memory Read changes only reward-time retrieval. Table~\ref{tab:harness_rounds} lists all searched attempts.}
  \label{fig:harness_revision_outcomes}
\end{figure}

\textbf{Effective revisions refine the feedback the policy learns from.}
All four retained revisions modify $H^{\mathrm{sig}}$ or $H^{\mathrm{opt}}$, and each refines the credit structure of its parent (Fig.~\ref{fig:harness_revision_outcomes}a).
$E_1$ replaces the session-level reward, shared by all operations, with per-operation credit measured by replaying the transaction without each operation.
$E_2$ measures that credit on the final bank instead of right after the session, and gives the largest gain: a write's value often shows only later, for example when a subsequent update overwrites what an earlier question needed.
$E_3$ groups credit around the operations that change memory state, and $E_4$ restricts variance control to operations whose credit is actually measured.
Read together, the sequence moves from one score per session toward credit for each decision, measured where its consequences appear.
Each step was proposed from the failures that remained after the previous revision, refining the credit structure progressively rather than specifying it at once.
After SC-GRPO saturates, the bottleneck is attributing answer outcomes to individual memory decisions.

\textbf{Revising the memory harness instead disrupts learning.}
To understand why the effective revisions concentrate on credit, we run the same recursive process restricted to $H^{\mathrm{mem}}$, in two settings (Fig.~\ref{fig:harness_revision_outcomes}b).
In \emph{Memory Build}, revisions change how the bank is structured, for example linking each record to more source rounds and images.
The score falls by 4.6 points: the new structure changes the states the policy had learned to handle, so it must relearn them, and the added links introduce redundant records.
In \emph{Memory Read}, revisions change the retrieval used to compute the training reward.
The reward rises during search, but the score falls by 1.0 point under the fixed retrieval of formal evaluation: better retrieval raised the reward without the policy writing a better bank.
Both failures share a cause.
Changing the memory structure alters the environment the policy acts in, and changing the reward-time retrieval alters what the reward measures; neither improves what the policy learns from its own decisions, which is what the credit revisions do.

\FloatBarrier

\section{Conclusion and Limitations}
\label{sec:conclusion}

\methodname{} introduces a framework in which a \role{memory manager} and the harness that trains it improve together.
A stateful Memory Environment organizes multimodal experience into State, History, and Records with source provenance, and Memory Policy Learning with SFT and SC-GRPO establishes a capable \role{manager} on it.
Recursive Policy--Harness Evolution then lets the \role{manager} diagnose its own memory experience and revise the harness it learns from.
\methodname{} achieves the highest score among controlled methods on three benchmarks and exceeds both frontier references, and recursive evolution brings further gains beyond policy learning under a fixed harness.
A single policy selected on Mem-Gallery generalizes to unseen benchmarks and \role{answer agents}, policy learning carries over to other \role{manager} backbones, and the learned memory is on average more compact than raw retrieval.
The analysis shows how successive revisions turn memory outcomes into more useful feedback for the \role{manager}'s next updates.

\paragraph{Limitations and future work.}
\methodname{} is developed primarily for personal assistance, in which a single \role{memory manager} maintains memory over multimodal conversations.
Other forms of agent memory are evaluated through transfer to web-agent trajectories, and settings such as embodied or multi-agent systems may involve additional forms of interaction and feedback in the harness.
The recursive evolution study uses 9B and 27B \role{managers} and Mem-Gallery as the
source benchmark. The retained revisions all change feedback and policy updates, while revisions to the memory harness did not help in our searches; other environments may favor different components. Future work could apply recursive evolution to other memory environments, such as procedural memory in web and embodied agents.
It can also scale the evolution search with larger budgets and more efficient strategies to explore more harness components.

{\small
\bibliographystyle{assets/plainnat}
\bibliography{references}

\begin{thebibliography}{81}
\providecommand{\natexlab}[1]{#1}
\providecommand{\url}[1]{\texttt{#1}}
\expandafter\ifx\csname urlstyle\endcsname\relax
  \providecommand{\doi}[1]{doi: #1}\else
  \providecommand{\doi}{doi: \begingroup \urlstyle{rm}\Url}\fi

\bibitem[Bei et~al.(2026)Bei, Wei, Ning, Zhao, Liu, Lin, Zhu, Hamann, He, and Tong]{bei2026memgallery}
Yuanchen Bei, Tianxin Wei, Xuying Ning, Yanjun Zhao, Zhining Liu, Xiao Lin, Yada Zhu, Hendrik Hamann, Jingrui He, and Hanghang Tong.
\newblock {Mem-Gallery}: Benchmarking multimodal long-term conversational memory for {MLLM} agents.
\newblock In \emph{Proceedings of the 64th Annual Meeting of the Association for Computational Linguistics (Volume 1: Long Papers)}, pages 40750--40784, San Diego, California, United States, July 2026. Association for Computational Linguistics.
\newblock \doi{10.18653/v1/2026.acl-long.1892}.
\newblock \url{https://aclanthology.org/2026.acl-long.1892/}.

\bibitem[Chen et~al.(2026)Chen, Jin, Helenowski, Yuan, Weston, Kulikov, Li, and Saha]{chen2026selfimprovingmemory}
Tong Chen, Chuanyang Jin, Eryk Helenowski, Weizhe Yuan, Jason Weston, Ilia Kulikov, Xian Li, and Swarnadeep Saha.
\newblock Self-improving memory: Meta-optimization and harness distillation for continual learning agents.
\newblock Blog post, RAM @ Meta AI, 2026.
\newblock \url{https://facebookresearch.github.io/RAM/blogs/self_improving_memory/}.

\bibitem[Chhikara et~al.(2025)Chhikara, Khant, Aryan, Singh, and Yadav]{chhikara2025mem0}
Prateek Chhikara, Dev Khant, Saket Aryan, Taranjeet Singh, and Deshraj Yadav.
\newblock {Mem0}: Building production-ready {AI} agents with scalable long-term memory.
\newblock \emph{arXiv preprint arXiv:2504.19413}, 2025.

\bibitem[{DeepSeek-AI}(2025)]{guo2025deepseekr1}
{DeepSeek-AI}.
\newblock {DeepSeek-R1}: Incentivizing reasoning capability in {LLMs} via reinforcement learning.
\newblock \emph{arXiv preprint arXiv:2501.12948}, 2025.

\bibitem[Du et~al.(2026)Du, Wang, Xiang, Wang, Huang, Xue, Liang, Zeng, Mi, Bai, Shang, Pan, Jiang, and Wong]{du2026memoryt1}
Yiming Du, Baojun Wang, Yifan Xiang, Zhaowei Wang, Wenyu Huang, Boyang Xue, Bin Liang, Xingshan Zeng, Fei Mi, Haoli Bai, Lifeng Shang, Jeff~Z. Pan, Yuxin Jiang, and Kam-Fai Wong.
\newblock {Memory-T1}: Reinforcement learning for temporal reasoning in multi-session agents.
\newblock In \emph{The Fourteenth International Conference on Learning Representations}, 2026.
\newblock arXiv:2512.20092.

\bibitem[Feng et~al.(2026)Feng, Xu, Chen, Dai, Wu, Li, Zeng, Xie, Liang, Lu, and Zhang]{feng2026m2a}
Junyu Feng, Binxiao Xu, Jiayi Chen, Mengyu Dai, Cenyang Wu, Haodong Li, Bohan Zeng, Yunliu Xie, Hao Liang, Ming Lu, and Wentao Zhang.
\newblock {M2A}: Multimodal memory agent with dual-layer hybrid memory for long-term personalized interactions, 2026.
\newblock \url{https://arxiv.org/abs/2602.07624}.

\bibitem[Feng et~al.(2025)Feng, Xue, Liu, and An]{feng2025gigpo}
Lang Feng, Zhenghai Xue, Tingcong Liu, and Bo~An.
\newblock Group-in-group policy optimization for {LLM} agent training.
\newblock In \emph{Advances in Neural Information Processing Systems}, 2025.
\newblock arXiv:2505.10978.

\bibitem[Fernando et~al.(2024)Fernando, Banarse, Michalewski, Osindero, and Rockt{\"a}schel]{fernando2023promptbreeder}
Chrisantha Fernando, Dylan Banarse, Henryk Michalewski, Simon Osindero, and Tim Rockt{\"a}schel.
\newblock Promptbreeder: Self-referential self-improvement via prompt evolution.
\newblock In \emph{International Conference on Machine Learning}, 2024.
\newblock arXiv:2309.16797.

\bibitem[Fisher(2025)]{fisher2025neural}
Matthew Fisher.
\newblock Neural graph memory: A structured approach to long-term memory in multimodal agents.
\newblock \emph{Zenodo preprint}, 2025.

\bibitem[Good(1966)]{good1966speculations}
Irving~John Good.
\newblock Speculations concerning the first ultraintelligent machine.
\newblock In \emph{Advances in Computers}, volume~6, pages 31--88. Academic Press, 1966.

\bibitem[Guo et~al.(2026)Guo, Jiao, Shi, Quan, Zhang, Li, Che, Xu, Liu, Liu, Kapadia, Pavlovic, Liu, Wang, Shi, Metaxas, and Tang]{guo2026memeye}
Minghao Guo, Qingyue Jiao, Zeru Shi, Yihao Quan, Boxuan Zhang, Danrui Li, Liwei Che, Wujiang Xu, Shilong Liu, Zirui Liu, Mubbasir Kapadia, Vladimir Pavlovic, Jiang Liu, Mengdi Wang, Yiyu Shi, Dimitris~N. Metaxas, and Ruixiang Tang.
\newblock {MemEye}: A visual-centric evaluation framework for multimodal agent memory, 2026.
\newblock \url{https://arxiv.org/abs/2605.15128}.

\bibitem[Hu et~al.(2025)Hu, Lu, and Clune]{hu2025adas}
Shengran Hu, Cong Lu, and Jeff Clune.
\newblock Automated design of agentic systems.
\newblock In \emph{International Conference on Learning Representations}, 2025.

\bibitem[Huang et~al.(2026)Huang, Zhang, Liang, Bei, Chen, Feng, Tan, Wang, Wei, Wu, et~al.]{huang2026survey}
Wei-Chieh Huang, Weizhi Zhang, Yueqing Liang, Yuanchen Bei, Yankai Chen, Tao Feng, Zhen Tan, Yu~Wang, Tianxin Wei, Shanglin Wu, et~al.
\newblock A survey of agent memory in the second half: Towards self-evolving and long-horizon agents.
\newblock \emph{Transactions on Machine Learning Research}, 2026.

\bibitem[Jaderberg et~al.(2017)Jaderberg, Dalibard, Osindero, Czarnecki, Donahue, Razavi, Vinyals, Green, Dunning, Simonyan, et~al.]{jaderberg2017pbt}
Max Jaderberg, Valentin Dalibard, Simon Osindero, Wojciech~M. Czarnecki, Jeff Donahue, Ali Razavi, Oriol Vinyals, Tim Green, Iain Dunning, Karen Simonyan, et~al.
\newblock Population based training of neural networks.
\newblock \emph{arXiv preprint arXiv:1711.09846}, 2017.

\bibitem[Jain et~al.(2025)Jain, Maheshwari, Yu, Hwu, and Shi]{jain2025augustus}
Jitesh Jain, Shubham Maheshwari, Ning Yu, Wen-mei Hwu, and Humphrey Shi.
\newblock Augustus: An llm-driven multimodal agent system with contextualized user memory.
\newblock \emph{arXiv preprint arXiv:2510.15261}, 2025.

\bibitem[Jiang et~al.(2025)Jiang, Hao, Cho, Li, Yuan, Chen, Ungar, Taylor, and Roth]{jiang2025know}
Bowen Jiang, Zhuoqun Hao, Young-Min Cho, Bryan Li, Yuan Yuan, Sihao Chen, Lyle Ungar, Camillo~J. Taylor, and Dan Roth.
\newblock Know me, respond to me: Benchmarking {LLM}s for dynamic user profiling and personalized responses at scale, 2025.
\newblock \url{https://arxiv.org/abs/2504.14225}.

\bibitem[Karten et~al.(2026)Karten, Zhang, Upaa~Jr, Feng, Li, Shi, Jin, and Vodrahalli]{karten2026continualharness}
Seth Karten, Joel Zhang, Tersoo Upaa~Jr, Ruirong Feng, Wenzhe Li, Chengshuai Shi, Chi Jin, and Kiran Vodrahalli.
\newblock Continual harness: Online adaptation for self-improving foundation agents.
\newblock \emph{arXiv preprint arXiv:2605.09998}, 2026.

\bibitem[Lee et~al.(2026{\natexlab{a}})Lee, Xu, Seely, Lee, Zaharia, and Tang]{lee2026recursiveharness}
Hyunin Lee, Jinglue Xu, Jeffrey Seely, Donghyun Lee, Matei Zaharia, and Yujin Tang.
\newblock Recursive harness self-improvement.
\newblock \emph{arXiv preprint arXiv:2607.15524}, 2026{\natexlab{a}}.
\newblock \url{https://arxiv.org/abs/2607.15524}.

\bibitem[Lee et~al.(2026{\natexlab{b}})Lee, Nair, Zhang, Lee, Khattab, and Finn]{lee2026metaharness}
Yoonho Lee, Roshen Nair, Qizheng Zhang, Kangwook Lee, Omar Khattab, and Chelsea Finn.
\newblock Meta-harness: End-to-end optimization of model harnesses.
\newblock \emph{arXiv preprint arXiv:2603.28052}, 2026{\natexlab{b}}.

\bibitem[Li et~al.(2025)Li, Zhang, Yang, Huang, Wu, Luo, Bei, Zou, Luo, Zhao, et~al.]{li2025survey}
Yangning Li, Weizhi Zhang, Yuyao Yang, Wei-Chieh Huang, Yaozu Wu, Junyu Luo, Yuanchen Bei, Henry~Peng Zou, Xiao Luo, Yusheng Zhao, et~al.
\newblock A survey of rag-reasoning systems in large language models.
\newblock In \emph{Findings of the Association for Computational Linguistics: EMNLP 2025}, pages 12120--12145, 2025.

\bibitem[Liu et~al.(2026{\natexlab{a}})Liu, Yang, Pu, Liu, Long, Guo, Chen, Weng, Kochkina, Kaur, Smiley, Liu, Zou, Liu, Bu, Peng, and Wang]{liu2026worldmemarena}
Chengzhi Liu, Yuzhe Yang, Sophia~Xiao Pu, Yepeng Liu, Lin Long, Yichen Guo, Nuo Chen, Zhaotian Weng, Elena Kochkina, Simerjot Kaur, Charese Smiley, Xiaomo Liu, James Zou, Sheng Liu, Yuheng Bu, Songyou Peng, and Xin~Eric Wang.
\newblock {WorldMemArena}: Evaluating multimodal agent memory through action-world interaction, 2026{\natexlab{a}}.
\newblock \url{https://arxiv.org/abs/2605.29341}.

\bibitem[Liu et~al.(2026{\natexlab{b}})Liu, Ye, Xia, Zheng, Xie, Ding, and Yao]{liu2026evolvemem}
Jiaqi Liu, Xinyu Ye, Peng Xia, Zeyu Zheng, Cihang Xie, Mingyu Ding, and Huaxiu Yao.
\newblock {EvolveMem}: Self-evolving memory architecture via {AutoResearch} for {LLM} agents.
\newblock \emph{arXiv preprint arXiv:2605.13941}, 2026{\natexlab{b}}.

\bibitem[Liu et~al.(2024)Liu, Lin, Hewitt, Paranjape, Bevilacqua, Petroni, and Liang]{liu2024lost}
Nelson~F. Liu, Kevin Lin, John Hewitt, Ashwin Paranjape, Michele Bevilacqua, Fabio Petroni, and Percy Liang.
\newblock Lost in the middle: How language models use long contexts.
\newblock \emph{Transactions of the Association for Computational Linguistics}, 12:\penalty0 157--173, 2024.
\newblock \doi{10.1162/tacl_a_00638}.
\newblock \url{https://aclanthology.org/2024.tacl-1.9/}.

\bibitem[Liu et~al.(2026{\natexlab{c}})Liu, Shi, Sang, He, Lin, Wei, Wang, Dumoulin, Jin, and Lu]{liu2026adaptiveautoharness}
Zewen Liu, Zhan Shi, Yisi Sang, Bing He, Minhua Lin, Tianxin Wei, Dakuo Wang, Benoit Dumoulin, Wei Jin, and Hanqing Lu.
\newblock Adaptive auto-harness: Sustained self-improvement for agentic system deployment on open-ended task streams.
\newblock \emph{arXiv preprint arXiv:2606.01770}, 2026{\natexlab{c}}.

\bibitem[Long et~al.(2025)Long, He, Ye, Pan, Lin, Li, Zhao, and Li]{long2025m3agent}
Lin Long, Yichen He, Wentao Ye, Yiyuan Pan, Yuan Lin, Hang Li, Junbo Zhao, and Wei Li.
\newblock Seeing, listening, remembering, and reasoning: A multimodal agent with long-term memory, 2025.
\newblock \url{https://arxiv.org/abs/2508.09736}.

\bibitem[Ma et~al.(2026)Ma, Feng, Huang, Feng, Ma, Xu, Gao, Hao, He, and Qin]{ma2026finemem}
Weitao Ma, Xiaocheng Feng, Lei Huang, Xiachong Feng, Zhanyu Ma, Jun Xu, Jiuchong Gao, Jinghua Hao, Renqing He, and Bing Qin.
\newblock {Fine-Mem}: Fine-grained feedback alignment for long-horizon memory management, 2026.
\newblock \url{https://arxiv.org/abs/2601.08435}.

\bibitem[Ma et~al.(2024)Ma, Liang, Wang, Huang, Bastani, Jayaraman, Zhu, Fan, and Anandkumar]{ma2024eureka}
Yecheng~Jason Ma, William Liang, Guanzhi Wang, De-An Huang, Osbert Bastani, Dinesh Jayaraman, Yuke Zhu, Linxi Fan, and Anima Anandkumar.
\newblock Eureka: Human-level reward design via coding large language models.
\newblock In \emph{The Twelfth International Conference on Learning Representations}, 2024.

\bibitem[Maharana et~al.(2024)Maharana, Lee, Tulyakov, Bansal, Barbieri, and Fang]{maharana2024locomo}
Adyasha Maharana, Dong-Ho Lee, Sergey Tulyakov, Mohit Bansal, Francesco Barbieri, and Yuwei Fang.
\newblock Evaluating very long-term conversational memory of {LLM} agents.
\newblock In \emph{Proceedings of the 62nd Annual Meeting of the Association for Computational Linguistics (Volume 1: Long Papers)}, pages 13851--13870, Bangkok, Thailand, August 2024. Association for Computational Linguistics.
\newblock \doi{10.18653/v1/2024.acl-long.747}.
\newblock \url{https://aclanthology.org/2024.acl-long.747/}.

\bibitem[Ning et~al.(2026)Ning, Fu, Wei, Bei, Yang, Xu, Song, Li, Li, Zeng, et~al.]{ning2026evoharness}
Xuying Ning, Dongqi Fu, Tianxin Wei, Yuanchen Bei, Xiyuan Yang, Wujiang Xu, Yueqi Song, Bingxuan Li, Zihao Li, Hanqing Zeng, et~al.
\newblock {EvoHarness-RL}: Learning runtime harness coordination for self-evolving agents.
\newblock \emph{arXiv preprint arXiv:2608.05446}, 2026.

\bibitem[{OpenAI}(2023)]{openai2023gpt4}
{OpenAI}.
\newblock {GPT-4} technical report.
\newblock \emph{arXiv preprint arXiv:2303.08774}, 2023.

\bibitem[Ouyang et~al.(2022)Ouyang, Wu, Jiang, Almeida, Wainwright, Mishkin, Zhang, Agarwal, Slama, Ray, et~al.]{ouyang2022training}
Long Ouyang, Jeff Wu, Xu~Jiang, Diogo Almeida, Carroll~L. Wainwright, Pamela Mishkin, Chong Zhang, Sandhini Agarwal, Katarina Slama, Alex Ray, et~al.
\newblock Training language models to follow instructions with human feedback.
\newblock In \emph{Advances in Neural Information Processing Systems}, volume~35, 2022.

\bibitem[Ouyang et~al.(2026)Ouyang, Yan, Hsu, Chen, Jiang, Wang, Han, Le, Daruki, Tang, Tirumalashetty, Lee, Rofouei, Lin, Han, Lee, and Pfister]{ouyang2026reasoningbank}
Siru Ouyang, Jun Yan, I-Hung Hsu, Yanfei Chen, Ke~Jiang, Zifeng Wang, Rujun Han, Long~T. Le, Samira Daruki, Xiangru Tang, Vishy Tirumalashetty, George Lee, Mahsan Rofouei, Hangfei Lin, Jiawei Han, Chen-Yu Lee, and Tomas Pfister.
\newblock {ReasoningBank}: Scaling agent self-evolving with reasoning memory.
\newblock In \emph{The Fourteenth International Conference on Learning Representations}, 2026.
\newblock arXiv:2509.25140.

\bibitem[Packer et~al.(2023)Packer, Wooders, Lin, Fang, Patil, Stoica, and Gonzalez]{packer2023memgpt}
Charles Packer, Sarah Wooders, Kevin Lin, Vivian Fang, Shishir~G. Patil, Ion Stoica, and Joseph~E. Gonzalez.
\newblock {MemGPT}: Towards {LLM}s as operating systems, 2023.
\newblock \url{https://arxiv.org/abs/2310.08560}.

\bibitem[Park et~al.(2023)Park, O'Brien, Cai, Morris, Liang, and Bernstein]{park2023generative}
Joon~Sung Park, Joseph O'Brien, Carrie~Jun Cai, Meredith~Ringel Morris, Percy Liang, and Michael~S Bernstein.
\newblock Generative agents: Interactive simulacra of human behavior.
\newblock In \emph{Proceedings of the 36th annual acm symposium on user interface software and technology}, pages 1--22, 2023.

\bibitem[Parker-Holder et~al.(2022)Parker-Holder, Rajan, Song, Biedenkapp, Miao, Eimer, Zhang, Nguyen, Calandra, Faust, Hutter, and Lindauer]{parkerholder2022autorl}
Jack Parker-Holder, Raghu Rajan, Xingyou Song, Andr{\'e} Biedenkapp, Yingjie Miao, Theresa Eimer, Baohe Zhang, Vu~Nguyen, Roberto Calandra, Aleksandra Faust, Frank Hutter, and Marius Lindauer.
\newblock Automated reinforcement learning ({AutoRL}): A survey and open problems.
\newblock \emph{Journal of Artificial Intelligence Research}, 74:\penalty0 517--568, 2022.

\bibitem[Ranzato et~al.(2016)Ranzato, Chopra, Auli, and Zaremba]{ranzato2016sequence}
Marc'Aurelio Ranzato, Sumit Chopra, Michael Auli, and Wojciech Zaremba.
\newblock Sequence level training with recurrent neural networks.
\newblock In \emph{International Conference on Learning Representations}, 2016.
\newblock arXiv:1511.06732.

\bibitem[Rasmussen et~al.(2025)Rasmussen, Paliychuk, Beauvais, Ryan, and Chalef]{rasmussen2025zep}
Preston Rasmussen, Pavlo Paliychuk, Travis Beauvais, Jack Ryan, and Daniel Chalef.
\newblock Zep: A temporal knowledge graph architecture for agent memory.
\newblock \emph{arXiv preprint arXiv:2501.13956}, 2025.

\bibitem[Ren et~al.(2026)Ren, Wang, Du, Xie, Liu, Yang, Feng, Pan, Zheng, Xu, et~al.]{ren2026memlens}
Xiyu Ren, Zhaowei Wang, Yiming Du, Zhongwei Xie, Chi Liu, Xinlin Yang, Haoyue Feng, Wenjun Pan, Tianshi Zheng, Baixuan Xu, et~al.
\newblock Memlens: Benchmarking multimodal long-term memory in large vision-language models.
\newblock \emph{arXiv preprint arXiv:2605.14906}, 2026.

\bibitem[Schmidhuber(2007)]{schmidhuber2007godel}
J{\"u}rgen Schmidhuber.
\newblock G{\"o}del machines: Fully self-referential optimal universal self-improvers.
\newblock In Ben Goertzel and Cassio Pennachin, editors, \emph{Artificial General Intelligence}, pages 199--226. Springer, 2007.

\bibitem[Shao et~al.(2024)Shao, Wang, Zhu, Xu, Song, Bi, Zhang, Zhang, Li, Wu, and Guo]{shao2024}
Zhihong Shao, Peiyi Wang, Qihao Zhu, Runxin Xu, Junxiao Song, Xiao Bi, Haowei Zhang, Mingchuan Zhang, Y.~K. Li, Y.~Wu, and Daya Guo.
\newblock {DeepSeekMath}: Pushing the limits of mathematical reasoning in open language models.
\newblock \emph{arXiv preprint arXiv:2402.03300}, 2024.
\newblock \url{https://arxiv.org/abs/2402.03300}.

\bibitem[Shen et~al.(2026)Shen, Wu, Lai, Lian, and Rao]{shen2026membuilder}
Zhiyu Shen, Ziming Wu, Fuming Lai, Shaobing Lian, and Yanghui Rao.
\newblock {MemBuilder}: Reinforcing {LLM}s for long-term memory construction via attributed dense rewards, 2026.
\newblock \url{https://arxiv.org/abs/2601.05488}.

\bibitem[Shinn et~al.(2023)Shinn, Cassano, Gopinath, Narasimhan, and Yao]{shinn2023reflexion}
Noah Shinn, Federico Cassano, Ashwin Gopinath, Karthik Narasimhan, and Shunyu Yao.
\newblock Reflexion: Language agents with verbal reinforcement learning.
\newblock In \emph{Advances in Neural Information Processing Systems}, volume~36, 2023.

\bibitem[Sun et~al.(2026)Sun, Lin, Xue, Wang, Yao, Qian, Xu, Li, Liu, Pan, Sun, Baker, and Zhou]{sun2026pmmc}
Jingyu Sun, Yan Lin, Yuyang Xue, Yifan Wang, Zhengtao Yao, Rui Qian, Zefeng Xu, Jiachen Li, Xianyang Liu, Jiancheng Pan, Jingyuan Sun, Syed~Murtuza Baker, and Hongpeng Zhou.
\newblock {PMMC}: Prospective multimodal memory compilation for long-term {LVLM} agents, 2026.
\newblock \url{https://arxiv.org/abs/2608.00962}.

\bibitem[Sutton and Barto(2018)]{sutton2018reinforcement}
Richard~S. Sutton and Andrew~G. Barto.
\newblock \emph{Reinforcement Learning: An Introduction}.
\newblock MIT Press, second edition, 2018.

\bibitem[Suzgun et~al.(2026)Suzgun, Yuksekgonul, Bianchi, Jurafsky, and Zou]{suzgun2026dc}
Mirac Suzgun, Mert Yuksekgonul, Federico Bianchi, Dan Jurafsky, and James Zou.
\newblock Dynamic cheatsheet: Test-time learning with adaptive memory.
\newblock In \emph{Proceedings of the 19th Conference of the European Chapter of the Association for Computational Linguistics (Volume 1: Long Papers)}, pages 7080--7106, 2026.

\bibitem[Wang et~al.(2026{\natexlab{a}})Wang, Lin, Wu, Zhang, Kang, Li, Qiu, Ye, Li, and Liu]{wang2026harnessrl}
Xiaoqiang Wang, Ting-En Lin, Yuchuan Wu, Pinyi Zhang, Caixin Kang, Ling Li, Bingxue Qiu, Jieping Ye, Yongbin Li, and Bang Liu.
\newblock Harnessrl: Co-evolving policies with a just-in-time training harness for open-ended rl.
\newblock \emph{arXiv preprint arXiv:2610.26426}, 2026{\natexlab{a}}.
\newblock \url{https://arxiv.org/abs/2610.26426}.

\bibitem[Wang and Chen(2025)]{wang2025mirix}
Yu~Wang and Xi~Chen.
\newblock {MIRIX}: Multi-agent memory system for {LLM}-based agents, 2025.
\newblock \url{https://arxiv.org/abs/2507.07957}.

\bibitem[Wang et~al.(2025)Wang, Takanobu, Liang, Mao, Hu, McAuley, and Wu]{wang2025memalpha}
Yu~Wang, Ryuichi Takanobu, Zhiqi Liang, Yuzhen Mao, Yuanzhe Hu, Julian McAuley, and Xiaojian Wu.
\newblock {Mem-$\alpha$}: Learning memory construction via reinforcement learning.
\newblock \emph{arXiv preprint arXiv:2509.25911}, 2025.

\bibitem[Wang et~al.(2026{\natexlab{b}})Wang, Zhao, Sun, Dong, Lan, Zhu, Li, Zheng, and Chen]{wang2026cmimem}
Yubo Wang, Qiuyu Zhao, Zenghui Sun, Shichao Dong, Jinsong Lan, Xiaoyong Zhu, Haoyang Li, Bo~Zheng, and Lei Chen.
\newblock {CMI-Mem}: Toward generalizable long-term memory management via {CMI}-augmented reinforcement learning, 2026{\natexlab{b}}.
\newblock \url{https://arxiv.org/abs/2607.20553}.

\bibitem[Wei et~al.(2025)Wei, Sachdeva, Coleman, He, Bei, Ning, Ai, Li, He, Chi, Wang, Chen, Pereira, Kang, and Cheng]{wei2025evomemory}
Tianxin Wei, Noveen Sachdeva, Benjamin Coleman, Zhankui He, Yuanchen Bei, Xuying Ning, Mengting Ai, Yunzhe Li, Jingrui He, Ed~H. Chi, Chi Wang, Shuo Chen, Fernando Pereira, Wang-Cheng Kang, and Derek~Zhiyuan Cheng.
\newblock {Evo-Memory}: Benchmarking {LLM} agent test-time learning with self-evolving memory.
\newblock \emph{arXiv preprint arXiv:2511.20857}, 2025.

\bibitem[Wei et~al.(2026{\natexlab{a}})Wei, Shi, Lin, He, Liu, Sang, Bei, Ning, Zou, Li, Lin, Zhao, Wang, Dumoulin, Wang, He, and Lu]{wei2026evo}
Tianxin Wei, Zhan Shi, Minhua Lin, Bing He, Zewen Liu, Yisi Sang, Yuanchen Bei, Xuying Ning, Jiaru Zou, Ting-Wei Li, Xiao Lin, Yanjun Zhao, Chi Wang, Benoit Dumoulin, Dakuo Wang, Jingrui He, and Hanqing Lu.
\newblock {Evo-Harness}: Context-to-harness skill compilation for self-evolving agents.
\newblock In \emph{Proceedings of the 2026 Conference on Empirical Methods in Natural Language Processing}, pages 22343--22358. Association for Computational Linguistics, October 2026{\natexlab{a}}.
\newblock \doi{10.18653/v1/2026.emnlp-main.969}.
\newblock \url{https://aclanthology.org/2026.emnlp-main.969/}.

\bibitem[Wei et~al.(2026{\natexlab{b}})Wei, Yao, Peng, Wang, Wu, Cheng, Meng, and Gao]{wei2026selfhpo}
Zhepei Wei, Wenlin Yao, Baolin Peng, Danqing Wang, Qianhui Wu, Hao Cheng, Yu~Meng, and Jianfeng Gao.
\newblock Self-evolving coding agents via harness-policy optimization.
\newblock \emph{arXiv preprint arXiv:2610.22791}, 2026{\natexlab{b}}.
\newblock \url{https://arxiv.org/abs/2610.22791}.

\bibitem[Wu et~al.(2025)Wu, Wang, Yu, Zhang, Chang, and Yu]{wu2025longmemeval}
Di~Wu, Hongwei Wang, Wenhao Yu, Yuwei Zhang, Kai-Wei Chang, and Dong Yu.
\newblock {LongMemEval}: Benchmarking chat assistants on long-term interactive memory.
\newblock In \emph{The Thirteenth International Conference on Learning Representations}, 2025.
\newblock \url{https://arxiv.org/abs/2410.10813}.

\bibitem[Wu et~al.(2026)Wu, Ji, Kawatkar, Kwan, Gu, Peng, and Chang]{wu2026longmemevalv2}
Di~Wu, Zixiang Ji, Asmi Kawatkar, Bryan Kwan, Jia-Chen Gu, Nanyun Peng, and Kai-Wei Chang.
\newblock {LongMemEval-V2}: Evaluating long-term agent memory toward experienced colleagues.
\newblock \emph{arXiv preprint arXiv:2605.12493}, 2026.

\bibitem[Xia et~al.(2026)Xia, Chen, Wang, Liu, Zeng, Wang, Han, Zhou, Zhao, Chen, et~al.]{xia2026skillrl}
Peng Xia, Jianwen Chen, Hanyang Wang, Jiaqi Liu, Kaide Zeng, Yu~Wang, Siwei Han, Yiyang Zhou, Xujiang Zhao, Haifeng Chen, et~al.
\newblock {SkillRL}: Evolving agents via recursive skill-augmented reinforcement learning.
\newblock \emph{arXiv preprint arXiv:2602.08234}, 2026.

\bibitem[Xiao et~al.(2026)Xiao, Lu, Wang, Zhang, Song, Wang, and Wang]{xiao2026reharness}
Yuxin Xiao, Zhiyun Lu, Jianyu Wang, Yangtian Zhang, Yang Song, Qifan Wang, and Bailin Wang.
\newblock Recurrent harness for agentic continual learning.
\newblock \emph{arXiv preprint arXiv:2610.12668}, 2026.

\bibitem[Xu et~al.(2025)Xu, Liang, Mei, Gao, Tan, and Zhang]{xu2025amem}
Wujiang Xu, Zujie Liang, Kai Mei, Hang Gao, Juntao Tan, and Yongfeng Zhang.
\newblock {A-Mem}: Agentic memory for {LLM} agents, 2025.
\newblock \url{https://arxiv.org/abs/2502.12110}.

\bibitem[Yan et~al.(2025)Yan, Yang, Huang, Nie, Ding, Li, Ma, Bi, Kersting, Pan, Sch{\"u}tze, Tresp, and Ma]{yan2025memoryr1}
Sikuan Yan, Xiufeng Yang, Zuchao Huang, Ercong Nie, Zifeng Ding, Zonggen Li, Xiaowen Ma, Jinhe Bi, Kristian Kersting, Jeff~Z. Pan, Hinrich Sch{\"u}tze, Volker Tresp, and Yunpu Ma.
\newblock {Memory-R1}: Enhancing large language model agents to manage and utilize memories via reinforcement learning, 2025.
\newblock \url{https://arxiv.org/abs/2508.19828}.

\bibitem[Yan et~al.(2026)Yan, Bahloul, Nie, Schwarzmann, Trivisonno, Tresp, and Ma]{yan2026memoryr2}
Sikuan Yan, Ahmed Bahloul, Ercong Nie, Susanna Schwarzmann, Riccardo Trivisonno, Volker Tresp, and Yunpu Ma.
\newblock {Memory-R2}: Fair credit assignment for long-horizon memory-augmented {LLM} agents, 2026.
\newblock \url{https://arxiv.org/abs/2605.21768}.

\bibitem[Yang et~al.(2025)Yang, Li, Yang, et~al.]{yang2025qwen3}
An~Yang, Anfeng Li, Baosong Yang, et~al.
\newblock {Qwen3} technical report.
\newblock \emph{arXiv preprint arXiv:2505.09388}, 2025.

\bibitem[Yang et~al.(2024{\natexlab{a}})Yang, Wang, Lu, Liu, Le, Zhou, and Chen]{yang2024opro}
Chengrun Yang, Xuezhi Wang, Yifeng Lu, Hanxiao Liu, Quoc~V. Le, Denny Zhou, and Xinyun Chen.
\newblock Large language models as optimizers.
\newblock In \emph{International Conference on Learning Representations}, 2024{\natexlab{a}}.

\bibitem[Yang et~al.(2024{\natexlab{b}})Yang, Jimenez, Wettig, Lieret, Yao, Narasimhan, and Press]{yang2024sweagent}
John Yang, Carlos~E. Jimenez, Alexander Wettig, Kilian Lieret, Shunyu Yao, Karthik Narasimhan, and Ofir Press.
\newblock {SWE-agent}: Agent-computer interfaces enable automated software engineering.
\newblock In \emph{Advances in Neural Information Processing Systems}, volume~37, 2024{\natexlab{b}}.

\bibitem[Yang et~al.(2026)Yang, Li, Shi, Li, Wang, Zhou, Mi, Yan, and Liang]{yang2026terminalrsi}
Junyao Yang, Zongxia Li, Yucheng Shi, Zhongzhi Li, Ruhan Wang, Ruiwen Zhou, Haitao Mi, Shuicheng Yan, and Leowei Liang.
\newblock {Terminal-RSI}: Recursive self-improving terminal agents via recursive self-rewrite and policy-refine.
\newblock \emph{arXiv preprint arXiv:2610.22192}, 2026.

\bibitem[Yeo et~al.(2026)Yeo, Kim, Jeong, Baek, and Hwang]{yeo2026universalrag}
Woongyeong Yeo, Kangsan Kim, Soyeong Jeong, Jinheon Baek, and Sung~Ju Hwang.
\newblock Universalrag: Retrieval-augmented generation over corpora of diverse modalities and granularities.
\newblock In \emph{Proceedings of the 64th Annual Meeting of the Association for Computational Linguistics (Volume 1: Long Papers)}, pages 3843--3871, 2026.

\bibitem[Yin et~al.(2025)Yin, Wang, Pan, Lin, Wan, and Wang]{yin2025godel}
Xunjian Yin, Xinyi Wang, Liangming Pan, Li~Lin, Xiaojun Wan, and William~Yang Wang.
\newblock G{\"o}del agent: A self-referential agent framework for recursively self-improvement.
\newblock In \emph{Proceedings of the 63rd Annual Meeting of the Association for Computational Linguistics (Volume 1: Long Papers)}, pages 27890--27913, 2025.
\newblock \doi{10.18653/v1/2025.acl-long.1354}.

\bibitem[Yu et~al.(2026{\natexlab{a}})Yu, Chen, Feng, Chen, Dai, Yu, Zhang, Ma, Liu, Wang, and Zhou]{yu2025memagent}
Hongli Yu, Tinghong Chen, Jiangtao Feng, Jiangjie Chen, Weinan Dai, Qiying Yu, Ya-Qin Zhang, Wei-Ying Ma, Jingjing Liu, Mingxuan Wang, and Hao Zhou.
\newblock {MemAgent}: Reshaping long-context {LLM} with multi-conv {RL}-based memory agent.
\newblock In \emph{International Conference on Learning Representations}, 2026{\natexlab{a}}.
\newblock arXiv:2507.02259.

\bibitem[Yu et~al.(2026{\natexlab{b}})Yu, Xia, Pan, Tang, Liu, Yu, Jiang, Chen, Xue, and Yu]{yu2026sim}
Xin Yu, Fei Xia, Jiaqi Pan, Youbao Tang, Chengzhi Liu, Adams Yu, Sicong Jiang, Yanfei Chen, Lingzhou Xue, and Ting Yu.
\newblock Self-improvement memory enables continual recursive self-improvement.
\newblock \emph{arXiv preprint arXiv:2611.18856}, 2026{\natexlab{b}}.
\newblock \url{https://arxiv.org/abs/2611.18856}.

\bibitem[Zelikman et~al.(2024)Zelikman, Lorch, Mackey, and Kalai]{zelikman2024stop}
Eric Zelikman, Eliana Lorch, Lester Mackey, and Adam~Tauman Kalai.
\newblock Self-taught optimizer ({STOP}): Recursively self-improving code generation.
\newblock In \emph{Conference on Language Modeling}, 2024.
\newblock arXiv:2310.02304.

\bibitem[Zhang et~al.(2026{\natexlab{a}})Zhang, Ren, Zhan, Wang, Zhu, Zhou, and Yan]{zhang2026memevolve}
Guibin Zhang, Haotian Ren, Chong Zhan, Junhao Wang, He~Zhu, Wangchunshu Zhou, and Shuicheng Yan.
\newblock {MemEvolve}: Meta-evolution of agent memory systems.
\newblock In \emph{Proceedings of the 43rd International Conference on Machine Learning}, volume 306 of \emph{Proceedings of Machine Learning Research}, pages 158215--158234, 2026{\natexlab{a}}.

\bibitem[Zhang et~al.(2026{\natexlab{b}})Zhang, Zhang, Li, Zhang, Chen, Zhang, Bai, and Hu]{zhang2026selfharness}
Hangfan Zhang, Shao Zhang, Kangcong Li, Chen Zhang, Yang Chen, Yiqun Zhang, Lei Bai, and Shuyue Hu.
\newblock Self-harness: Harnesses that improve themselves.
\newblock \emph{arXiv preprint arXiv:2606.09498}, 2026{\natexlab{b}}.

\bibitem[Zhang et~al.(2026{\natexlab{c}})Zhang, Long, Bao, Feng, Zhang, Yue, and Wang]{zhang2026memskill}
Haozhen Zhang, Quanyu Long, Jianzhu Bao, Tao Feng, Weizhi Zhang, Haodong Yue, and Wenya Wang.
\newblock {MemSkill}: Learning and evolving memory skills for self-evolving agents.
\newblock In \emph{Advances in Neural Information Processing Systems}, 2026{\natexlab{c}}.
\newblock arXiv:2602.02474.

\bibitem[Zhang et~al.(2026{\natexlab{d}})Zhang, Hu, Lu, Lange, and Clune]{zhang2025dgm}
Jenny Zhang, Shengran Hu, Cong Lu, Robert Lange, and Jeff Clune.
\newblock {Darwin G{\"o}del Machine}: Open-ended evolution of self-improving agents.
\newblock In \emph{The Fourteenth International Conference on Learning Representations}, 2026{\natexlab{d}}.
\newblock \url{https://arxiv.org/abs/2505.22954}.

\bibitem[Zhang et~al.(2025{\natexlab{a}})Zhang, Hu, Upasani, Ma, Hong, Kamanuru, Rainton, Wu, Ji, Li, et~al.]{zhang2025ace}
Qizheng Zhang, Changran Hu, Shubhangi Upasani, Boyuan Ma, Fenglu Hong, Vamsidhar Kamanuru, Jay Rainton, Chen Wu, Mengmeng Ji, Hanchen Li, et~al.
\newblock Agentic context engineering: Evolving contexts for self-improving language models.
\newblock \emph{arXiv preprint arXiv:2510.04618}, 2025{\natexlab{a}}.

\bibitem[Zhang et~al.(2025{\natexlab{b}})Zhang, Zhang, Xie, Li, Dai, Long, Xie, Zhang, Li, and Zhang]{zhang2024gme}
Xin Zhang, Yanzhao Zhang, Wen Xie, Mingxin Li, Ziqi Dai, Dingkun Long, Pengjun Xie, Meishan Zhang, Wenjie Li, and Min Zhang.
\newblock Bridging modalities: Improving universal multimodal retrieval by multimodal large language models.
\newblock In \emph{Proceedings of the IEEE/CVF Conference on Computer Vision and Pattern Recognition (CVPR)}, pages 9274--9285, 2025{\natexlab{b}}.
\newblock arXiv:2412.16855.

\bibitem[Zhao et~al.(2024)Zhao, Huang, Xu, Lin, Liu, and Huang]{zhao2024expel}
Andrew Zhao, Daniel Huang, Quentin Xu, Matthieu Lin, Yong-Jin Liu, and Gao Huang.
\newblock {ExpeL}: {LLM} agents are experiential learners.
\newblock In \emph{Proceedings of the AAAI Conference on Artificial Intelligence}, volume~38, pages 19632--19642, 2024.

\bibitem[Zheng et~al.(2026)Zheng, Shi, Cai, Li, Zhang, Li, Yu, and Ma]{zheng2026lifelong}
Junhao Zheng, Chengming Shi, Xidi Cai, Qiuke Li, Duzhen Zhang, Chenxing Li, Dong Yu, and Qianli Ma.
\newblock Lifelong learning of large language model based agents: A roadmap.
\newblock \emph{IEEE Transactions on Pattern Analysis and Machine Intelligence}, 2026.
\newblock \doi{10.1109/TPAMI.2025.3650546}.

\bibitem[Zhong et~al.(2024)Zhong, Guo, Gao, Ye, and Wang]{zhong2024memorybank}
Wanjun Zhong, Lianghong Guo, Qiqi Gao, He~Ye, and Yanlin Wang.
\newblock {MemoryBank}: Enhancing large language models with long-term memory.
\newblock In \emph{Proceedings of the AAAI Conference on Artificial Intelligence}, volume~38, 2024.

\bibitem[Zhou et~al.(2024{\natexlab{a}})Zhou, Xu, Zhu, Zhou, Lo, Sridhar, Cheng, Ou, Bisk, Fried, Alon, and Neubig]{zhou2024webarena}
Shuyan Zhou, Frank~F. Xu, Hao Zhu, Xuhui Zhou, Robert Lo, Abishek Sridhar, Xianyi Cheng, Tianyue Ou, Yonatan Bisk, Daniel Fried, Uri Alon, and Graham Neubig.
\newblock {WebArena}: A realistic web environment for building autonomous agents.
\newblock In \emph{The Twelfth International Conference on Learning Representations}, 2024{\natexlab{a}}.

\bibitem[Zhou et~al.(2024{\natexlab{b}})Zhou, Zanette, Pan, Levine, and Kumar]{zhou2024archer}
Yifei Zhou, Andrea Zanette, Jiayi Pan, Sergey Levine, and Aviral Kumar.
\newblock {ArCHer}: Training language model agents via hierarchical multi-turn {RL}.
\newblock In \emph{International Conference on Machine Learning}, 2024{\natexlab{b}}.

\bibitem[Zhu et~al.(2026)Zhu, Chen, Yu, Wu, and Wang]{zhu2026tiermem}
Qiming Zhu, Shunian Chen, Rui Yu, Zhehao Wu, and Benyou Wang.
\newblock From lossy to verified: A provenance-aware tiered memory for agents, 2026.
\newblock \url{https://arxiv.org/abs/2602.17913}.

\bibitem[Zou et~al.(2026)Zou, He, Qiu, Lin, and Li]{zou2026taskmem}
Tao Zou, Yichen He, Tian Qiu, Yuan Lin, and Hang Li.
\newblock Task-focused memorization for multimodal agents, 2026.
\newblock \url{https://arxiv.org/abs/2605.31075}.

\end{thebibliography}
}

\appendix
\raggedbottom

\titlecontents{lsection}[1.6em]{\vspace{6pt}\bfseries}
  {\contentslabel{1.6em}}{}{\hfill\contentspage}
\titlecontents{lsubsection}[4.2em]{\vspace{2pt}}
  {\contentslabel{2.6em}}{}{\titlerule*[0.7pc]{.}\contentspage}
{\sffamily\bfseries\large Table of Contents\par}
\vspace{4pt}
\startcontents[appendix]
\printcontents[appendix]{l}{1}[2]{}
\vspace{12pt}

\section{Implementation Details}
\label{app:reproducibility_controls}

\subsection{Benchmarks and Baselines}
\label{app:benchmarks_baselines}

\paragraph{Benchmark coverage.}
MemLens uses the pinned agent subset of 195 questions, 2,613 sessions, and 2,494 resolved images.
LongMemEval-cleaned contains 500 questions, and LongMemEval-v2 keeps its browser screenshots and question images.

\paragraph{Baseline adaptation.}
All training-free baselines share the dataset adapter and \role{answer agent} while keeping their own storage and retrieval mechanisms.
Memory-R2 uses its public checkpoint and prompt, constrained JSON decoding, and the published no-op fallback, and MemAgent is reported as a second trained \role{manager}.
MemBuilder~\citep{shen2026membuilder} and Memory-T1~\citep{du2026memoryt1} are omitted because complete public weights are unavailable.

\subsection{Inference and Scoring}
\label{app:inference_scoring}

At evaluation, \role{manager} transactions are decoded greedily with thinking disabled (temperature 0, top-$p$ 1, seed 42), with up to 229,376 input tokens and 32,768 output tokens in a 262,144-token engine context.
Write and update retrieval uses Qwen3-VL-Embedding-2B.
For answering, each read path is frozen: evidence is ranked by cosine similarity with \texttt{gme-Qwen2-VL-2B-Instruct}, ten units are selected, and their original source rounds are merged in order without duplicates.
Unless noted otherwise, the \role{answer agent} is BF16 Qwen3.5-9B with medium thinking effort, a 32,768-token context, a 4,096-token output cap, and temperature 0; PersonaMem uses the same context limit.
Empty or truncated answers are retried once with thinking disabled and a 256-token cap.
The judge score averages Llama-3.3 grades, and rounded accuracy counts grades of at least 0.5.
Benchmark-specific token F1 is preserved, and PersonaMem uses normalized MCQ exact match.

\subsection{Policy Learning}
\label{app:policy_learning}

\paragraph{Policy initialization.}
Gemini 3.1 Pro generates demonstration transactions from each session, the current bank, and the available operations.
Mem-Gallery SFT trains for three epochs on 69 audited transactions from six dialogues, with a seventh held out, using LoRA rank 16 and scale 32 and excluding prompt tokens from the loss.
RL uses 181 sessions and 1,091 grounded questions from 15 training dialogues.
A question is scored after its last supporting clue appears and stays hidden from the \role{manager}.
Each session receives eight rollouts at temperature 1, all starting from the same historical state and each executed on its own bank copy.
Prefix banks are generated greedily and refreshed between epochs.

\paragraph{Outcome learning.}
The reward combines answer F1 (0.85) with category-balanced answer quality (0.15) and penalizes truncation, duplicate actions, missing EOS, excessive length, invalid image references, and missing answer entities.
A validity gate blocks positive advantage for recovered invalid writes.
Each update uses eight groups of eight rollouts, AdamW at $10^{-5}$, gradient clipping at 1, and KL weight $\beta=0.03$ against the policy at the start of the round, with \role{manager} rollouts capped at 8,192 tokens.
Training rewards use deterministic answers with thinking disabled and a 256-token cap, while formal evaluation uses the frozen protocol above.

\subsection{Recursive Evolution Protocol}
\label{app:evolution_protocol}

\paragraph{Selection protocol.}
Because of the computational budget, the 9B search allows at most ten rejected revisions ($A_1$--$A_{10}$) and stops early when this limit is reached; by then, four revisions have been retained ($E_1$--$E_4$; Table~\ref{tab:harness_rounds}).
Each 9B candidate starts from its recorded parent.
A completed run uses ten updates and 640 sampled trajectories on the same episode schedule, with learning rates of $10^{-5}$ and $2\times10^{-6}$ for the first two retained revisions and $10^{-6}$ for later searches.
The BF16 Qwen3.5-9B proposer receives the current harness, training outcomes, diagnostic evidence, and protocol constraints, and returns one revision.
Revisions are selected on 426 questions from four conversations held out from policy training; this development set is reused across candidates.
GME Top-10 retrieval, merged original-round evidence, the thinking 9B \role{answer agent}, and the Llama-3.3 judge stay fixed.
The continuation control matches the update count of the retained path, excluding evaluations of other proposals.

\paragraph{Credit revisions and implementation checks.}
The retained revisions (\S\ref{sec:recursive_harness_adaptation}) all redistribute credit within a transaction.
Such a revision splits the actions $a_{1:n}$ of one transaction into spans with a segmentation rule $\operatorname{Seg}_H$ and estimates the contribution $\Delta_i$ of each span by bounded executor replay.
A credit router $C_H$ then redistributes the session-level group advantage $A_g$ of Eq.~\ref{eq:group-advantage} into span-level advantages $\widetilde A_{1:n}$, subject to conservation:
\begin{equation}
\begin{aligned}
\widetilde A_{1:n}
 &=C_H\!\left(A_g,\operatorname{Seg}_H(a_{1:n}),\Delta_{1:n}\right),\\
\sum_i\widetilde A_i&=A_g.
\end{aligned}
\label{eq:evolved_credit_router}
\end{equation}
Both $\operatorname{Seg}_H$ and $C_H$ are part of the harness and can change across revisions.
Spans without replay support receive zero credit.
Before training, each revision is checked for activation, finite values, the support mask, and conservation of credit, and it is selected by memory utility after training.

\paragraph{Audit trail.}
Proposals, diffs, training settings, checkpoints, and available evaluations are retained.
Coding-assisted repairs to trace alignment and to numerical or conservation guards are recorded separately.

\FloatBarrier
\section{Additional Results}
\label{app:additional_results}

\subsection{Memory Quality (RQ1)}
\label{app:long_horizon_efficiency}

\begin{table}[!t]
\centering
\scriptsize
\setlength{\tabcolsep}{3.6pt}
\renewcommand{\arraystretch}{1.08}
\begin{tabular}{llcccccccccc}
\toprule
\textbf{Backbone} & \textbf{Stage} & \textbf{FR} & \textbf{VS} & \textbf{TTL} &
\textbf{TR} & \textbf{VR} & \textbf{MR} & \textbf{KR} & \textbf{CD} & \textbf{AR} & \textbf{Overall} \\
\midrule
\multirow{3}{*}{Qwen3.5-9B}
& \envshort{} & 81.5 & 78.4 & \textbf{90.8} & 83.3 & 80.3 & 73.6 & \textbf{84.7} & \textbf{100.0} & \textbf{90.5} & 83.8 \\
& \warmupshort{} & 83.7 & 84.2 & 88.1 & \textbf{100.0} & 70.2 & 77.7 & 69.4 & \textbf{100.0} & 84.5 & 83.2 \\
& \methodname{} & \textbf{84.2} & \textbf{90.3} & 90.5 & \textbf{100.0} & \textbf{82.7} & \textbf{82.4} & 83.3 & \textbf{100.0} & 86.9 & \textbf{88.0} \\
\bottomrule
\end{tabular}
\caption{\textbf{Mem-Gallery results by task category.} Judge scores
(\%) under the fixed read path. Categories cover memory extraction and
adaptation (FR: factual retrieval, VS: visual-centric search, TTL: test-time
learning), memory reasoning (TR: temporal reasoning, VR: visual-centric
reasoning, MR: multi-entity reasoning), and knowledge management (KR: knowledge
resolution, CD: conflict detection, AR: answer refusal)
\citep{bei2026memgallery}.}
\label{tab:posttraining}
\end{table}

\paragraph{Task categories.}
Table~\ref{tab:posttraining} breaks down Mem-Gallery by task category.
The largest gains over \envname{} appear in temporal reasoning, visual-centric search, and multi-entity reasoning.
These categories require locating a specific observation or combining evidence across sessions, consistent with the main finding that post-training helps most on multi-step and temporal questions.
Answer refusal is the only category with a clear drop (90.5 to 86.9): with a richer bank, the \role{answer agent} more often finds related evidence and answers questions it should decline (\S\ref{sec:main_results}).

\begin{figure}[!t]
  \centering
  \includegraphics[width=0.94\textwidth]{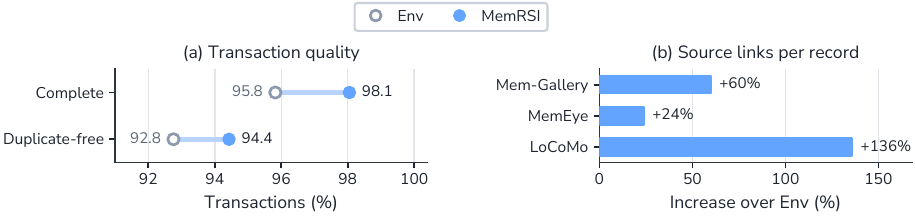}
  \caption{\textbf{Write quality and source links.} (a) Share of the 359 session transactions, pooled over the three benchmarks, that are complete (not truncated) and duplicate-free (no duplicate operations). (b) Increase over \envname{} in source rounds linked per record. Policies are trained on Mem-Gallery.}
  \label{fig:policy_behavior}
\end{figure}

\paragraph{Write quality.}
After training, more transactions are complete and free of duplicates, and each record links to more source rounds (Fig.~\ref{fig:policy_behavior}).
The trained \role{manager} therefore writes updates that execute as intended and keeps the evidence links that answering later relies on.

\begin{figure}[!t]
  \centering
  \includegraphics[width=\textwidth]{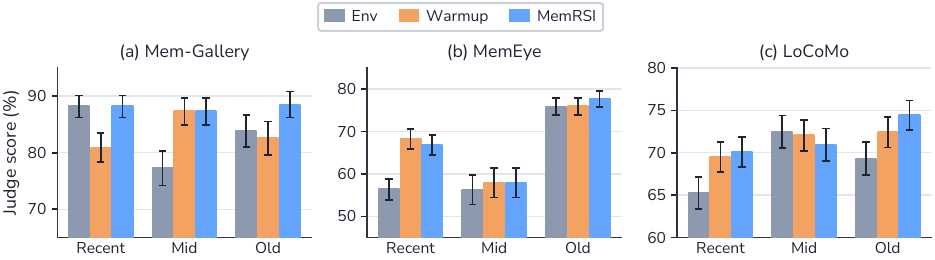}
  \vspace{2pt}
  \includegraphics[width=\textwidth]{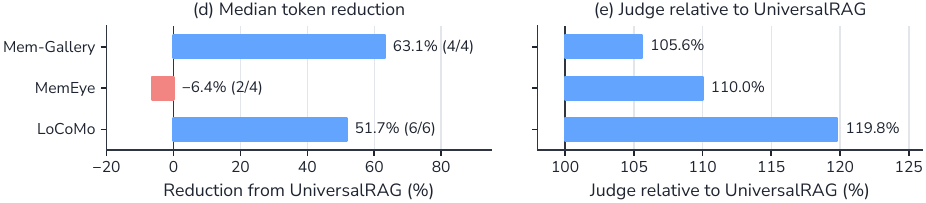}
  \caption{\textbf{Evidence age and memory efficiency} (Mem-Gallery-trained policy). (a--c) Mean judge and standard error by age of the latest source round. (d) Median per-history token reduction (48.4\% over all 14 histories) and (e) relative judge against UniversalRAG, where 100\% means equal quality. Parentheses count histories with smaller banks.}
  \label{fig:long_horizon_quality}
  \label{fig:memory_efficiency}
\end{figure}

\paragraph{Memory efficiency and evidence age.}
Compared with UniversalRAG, the learned banks are smaller in 12 of 14 histories and reach higher judge scores at the dataset level (Fig.~\ref{fig:memory_efficiency}).
On MemEye, bank size relative to UniversalRAG varies across histories, yet the relative judge stays about 10\% higher, indicating that the gain comes from what the bank stores rather than from storing more.
Evidence-age bins are assigned after inference and leave retrieval unchanged.
The learned policy gains on questions whose evidence appeared long before, showing that early observations remain accessible; since each bin holds a different group of questions, the bins compare question groups rather than measure memory decay (Fig.~\ref{fig:long_horizon_quality}).

\FloatBarrier
\subsection{Generalization and Source Evidence (RQ2)}
\label{app:cross_dataset_cases}
\begin{figure}[!t]
  \centering
  \includegraphics[width=0.74\linewidth]{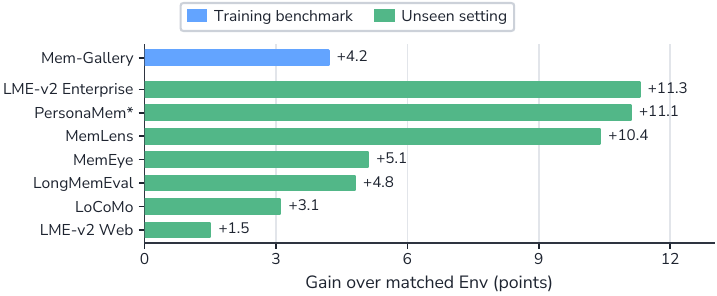}
  \caption{\textbf{Gains of the Mem-Gallery policy over \envname{}} (points). Blue: training benchmark; green: seven unseen settings. *PersonaMem uses normalized MCQ exact match; the other settings use the judge score.}
  \label{fig:cross_dataset_generalization}
\end{figure}

\paragraph{Transfer gains.}
The Mem-Gallery policy is reused on seven unseen settings and compared with the matching \envname{} \role{manager} in each (Fig.~\ref{fig:cross_dataset_generalization}).
Every setting improves, and the largest gains appear on MemLens, LongMemEval-v2 Enterprise, and PersonaMem, whose histories are long or heterogeneous.
This suggests that the learned construction policy matters most when the raw history is hardest to organize.

\begin{figure}[!t]
  \centering
  \includegraphics[width=0.97\textwidth]{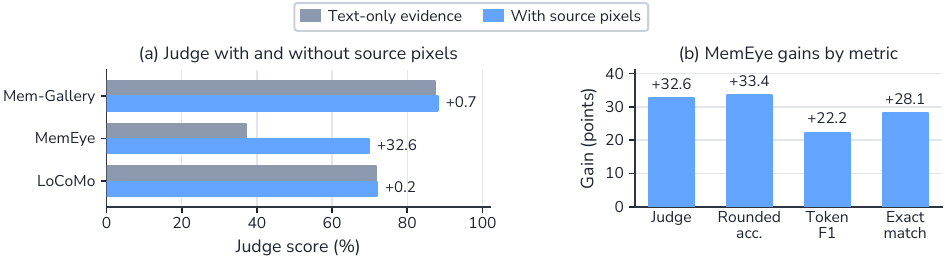}
  \caption{\textbf{Contribution of source images.} (a) Judge scores with and without linked pixels; memory text, captions, and retrieval are unchanged. (b) MemEye gains in judge score, rounded accuracy (grades of at least 0.5), token F1, and exact match. Memory comes from the Mem-Gallery-trained policy.}
  \label{fig:multimodal_contribution}
\end{figure}

\paragraph{Visual evidence.}
With memory text, captions, and retrieval rankings fixed, only access to the linked source pixels is varied (Fig.~\ref{fig:multimodal_contribution}).
Restoring the pixels adds 32.6 judge points on MemEye, while Mem-Gallery and LoCoMo change little.
MemEye questions depend on fine visual details that captions and compressed descriptions do not recover, so the source links let the \role{manager} consolidate memory without losing evidence that an abstraction may omit.

\FloatBarrier
\subsection{Recursive Evolution (RQ3)}
\label{app:learning_harness_evolution}

\begin{table}[!t]
\centering\small
\setlength{\tabcolsep}{5pt}
\begin{tabularx}{\textwidth}{Xlrrr}
\toprule
\textbf{Policy} & \textbf{Source decision} & \textbf{Mem-Gallery} & \textbf{MemEye} & \textbf{LoCoMo} \\
\midrule
\initname{} & Starting point & 85.62 & 66.53 & \textbf{71.98} \\
\rowcolor{YC} \methodname{} ($E_4$) & Retained & \textbf{88.03} & 69.73 & 71.87 \\
Ordered-prefix candidate (A9) & Not retained & 85.74 & 68.59 & 71.59 \\
Trunk/branch candidate (A10) & Not retained & 87.21 & \textbf{70.70} & 71.12 \\
\bottomrule
\end{tabularx}
\caption{\textbf{Transfer of candidate revisions.}
Judge scores (\%) with the 9B answer agent. $E_4$ is selected on Mem-Gallery;
A9 and A10 both start from $E_4$. MemEye (1,030 questions) and LoCoMo (1,174)
serve as transfer settings outside selection.}
\label{tab:learning_harness_transfer}
\end{table}

\begin{table}[!t]
\centering\small
\setlength{\tabcolsep}{6pt}
\begin{tabularx}{\textwidth}{Xrrrr}
\toprule
\textbf{Benchmark} & \textbf{\envname{}} & \textbf{\initname{}} & \textbf{\methodname} & \textbf{Change} \\
\midrule
Mem-Gallery (source) & 83.8 & 85.6 & \textbf{88.0} & $+2.4$ \\
\midrule
\multicolumn{5}{l}{\textit{Unseen settings}} \\
MemEye & 64.6 & 66.5 & \textbf{69.7} & $+3.2$ \\
MemLens & 52.7 & 62.4 & \textbf{63.1} & $+0.7$ \\
LME-v2 Web & 52.4 & 50.1 & \textbf{53.9} & $+3.8$ \\
LME-v2 Enterprise & 41.4 & 51.0 & \textbf{52.7} & $+1.7$ \\
PersonaMem$^\dagger$ & 61.6 & 70.5 & \textbf{72.7} & $+2.2$ \\
LoCoMo & 68.8 & \textbf{72.0} & 71.9 & $-0.1$ \\
LongMemEval & 76.8 & \textbf{84.2} & 81.6 & $-2.6$ \\
\bottomrule
\end{tabularx}
\caption{\textbf{\envname{}, \initname{}, and \methodname{} on the source and unseen settings.}
Both trained policies use Mem-Gallery only, and the seven unseen settings are excluded
from proposal and selection. MemEye and LoCoMo scores therefore differ from
Table~\ref{tab:main_results}, which reports the policy trained on each benchmark. Change is \methodname{} minus \initname{}.
$^\dagger$PersonaMem uses normalized MCQ exact match; the other rows use the judge score.}
\label{tab:cross_dataset_grpo}
\end{table}

\paragraph{Transfer of revisions.}
Table~\ref{tab:cross_dataset_grpo} follows each setting from the Memory Environment through SC-GRPO to recursive revision.
SC-GRPO training on Mem-Gallery lifts most settings above \envname{}, recursive revision adds further gains in five of the seven unseen settings, and with both, \methodname{} is above \envname{} on all eight.
Revision is driven only by Mem-Gallery, which is multimodal, and its gains transfer best to the other multimodal settings: MemEye and LongMemEval-v2 Web, which keeps its browser screenshots, gain the most.
The two settings without a revision gain, LoCoMo and LongMemEval, are conversational benchmarks whose questions rest mainly on text: LongMemEval has no images, and removing the source pixels barely changes LoCoMo (Fig.~\ref{fig:multimodal_contribution}).
Two rejected candidates show that selection on one benchmark does not always pick the best revision for another (Table~\ref{tab:learning_harness_transfer}).
A10 falls short of $E_4$ on Mem-Gallery and is therefore not retained, but it scores highest on MemEye; A9 is below $E_4$ on all three benchmarks.

\paragraph{Revision record.}
Table~\ref{tab:harness_rounds} classifies each 9B attempt by what it changes relative to its parent: a signal revision changes what is scored, and an optimization revision changes how that feedback is grouped or weighted.
Attempts that end before scoring are classified by their proposal.
\begin{table}[!t]
\centering\footnotesize
\setlength{\tabcolsep}{4pt}
\renewcommand{\arraystretch}{1.12}
\begin{tabularx}{\linewidth}{ll>{$}l<{$}>{\raggedright\arraybackslash}Xrrl}
\toprule
\textbf{Attempt} & \textbf{Parent} & \multicolumn{1}{l}{\textbf{Component}} & \textbf{Revision} & \textbf{Judge} & \textbf{Change} & \textbf{Decision} \\
\midrule
$E_{0}$ & --- & \text{---} & \initname{} & 85.62 & --- & Start \\
$A_{1}$ & $E_{0}$ & H^{\mathrm{sig}}+H^{\mathrm{opt}} & Provenance-filtered action deletion with group centering & --- & --- & Stopped at 2/10 \\
$E_{1}$ & $E_{0}$ & H^{\mathrm{sig}}+H^{\mathrm{opt}} & Leave-one-action-out replay and action-level credit & 85.92 & $+0.30$ & \textbf{Retained} \\
$A_{2}$ & $E_{0}$ & H^{\mathrm{opt}} & Independent fixed five-turn chunks & 84.15 & $-1.47$ & Not retained \\
$A_{3}$ & $E_{1}$ & H^{\mathrm{opt}} & Stateful fixed five-turn chunks & 84.86 & $-1.06$ & Not retained \\
$E_{2}$ & $E_{1}$ & H^{\mathrm{sig}} & Measure action contributions at the final bank & 86.97 & $+1.05$ & \textbf{Retained} \\
$A_{4}$ & $E_{2}$ & H^{\mathrm{sig}} & Select replay counterfactuals by operation type & 85.33 & $-1.64$ & Not retained \\
$E_{3}$ & $E_{2}$ & H^{\mathrm{opt}} & Group credit at state-changing operations & 87.21 & $+0.24$ & \textbf{Retained} \\
$A_{5}$ & $E_{3}$ & H^{\mathrm{opt}} & Weight segment credit by action density & 83.57 & $-3.64$ & Not retained \\
$A_{6}$ & $E_{3}$ & H^{\mathrm{opt}} & Shrink credit by confidence & 86.74 & $-0.47$ & Not retained \\
$E_{4}$ & $E_{3}$ & H^{\mathrm{opt}} & Restrict variance control to supported actions & 88.03 & $+0.82$ & \textbf{Retained} \\
$A_{7}$ & $E_{4}$ & H^{\mathrm{sig}}+H^{\mathrm{opt}} & Measure per-QA sign and coherence to route credit & 85.09 & $-2.94$ & Not retained \\
$A_{8}$ & $E_{4}$ & H^{\mathrm{sig}} & Estimate joint contributions of actions by coalition & --- & --- & Inactive \\
$A_{9}$ & $E_{4}$ & H^{\mathrm{sig}}+H^{\mathrm{opt}} & Score ordered-prefix deletions and assign path credit & 85.74 & $-2.29$ & Not retained \\
$A_{10}$ & $E_{4}$ & H^{\mathrm{opt}} & Route credit by state-transition trunk and branch & 87.21 & $-0.82$ & Not retained \\
\bottomrule
\end{tabularx}
\caption{\textbf{9B revision attempts on Mem-Gallery.}
Judge is reported as a percentage with the 9B answer agent; change is in points
relative to the listed parent. $A_1$ stopped during training; $A_8$ ended at
the activation check before the first update.}
\label{tab:harness_rounds}
\end{table}

\paragraph{Memory-side search diagnostics.}
Construction revisions expanded coverage of source rounds and images through provenance closure, which also introduced redundant records.
Retrieval revisions changed the evidence supplied to answer calls during reward computation, and their local search gains did not persist in formal QA (Fig.~\ref{fig:harness_revision_outcomes}b).
Formal evaluation keeps frozen GME Top-10 retrieval and merged original-round evidence throughout these searches.

\FloatBarrier
\section{Case Studies}
\label{app:case_studies}
\begin{figure}[tp]
  \centering
  \includegraphics[width=\textwidth]{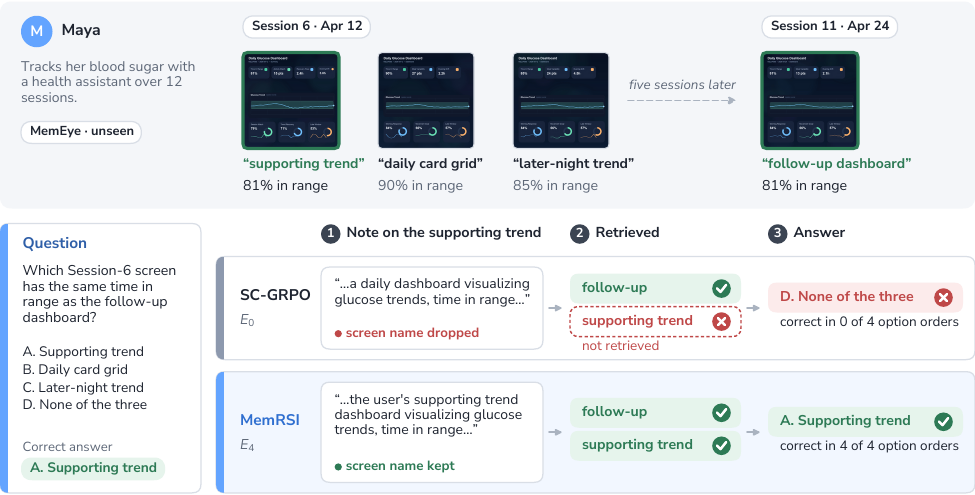}
  \caption{\textbf{Evidence recovery on an unseen MemEye history.} Top: Maya shares three similar glucose dashboards in Session~6 and a follow-up five sessions later; labels are her names for the screens, and green frames mark the two with the same time in range. Bottom: the question, whose options name the Session-6 screens and are asked in four orders, and for each \role{manager}, the note stored for the supporting trend (1), the matching screens retrieval returns in every order (2), and the answer (3). \envname{} fails in all four orders.}
  \label{fig:stagewise_multimodal_case}
\end{figure}

\begin{figure}[tp]
  \centering
  \includegraphics[width=\linewidth]{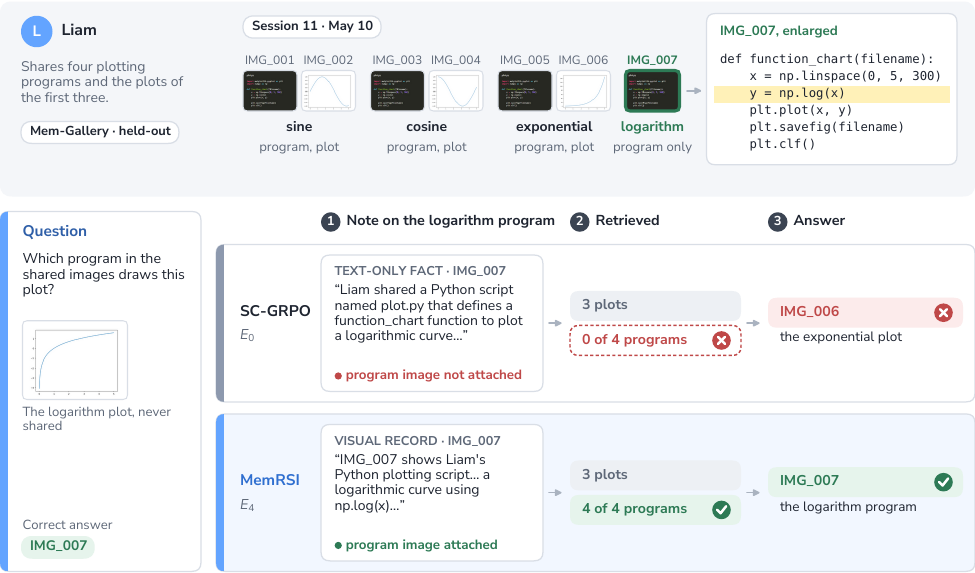}
  \caption{\textbf{Program recovery on a held-out Mem-Gallery history.} Top: the program and plot images Liam shares in Session~11, with the logarithm program enlarged; its plot is never shared. Bottom: for each \role{manager}, the record type and note stored for the logarithm program (1), what retrieval returns for the question, namely the three plots and how many of the four programs (2), and the answer (3).}
  \label{fig:memory_quality_cases}
\end{figure}

\begin{figure}[t]
  \centering
  \includegraphics[width=\linewidth]{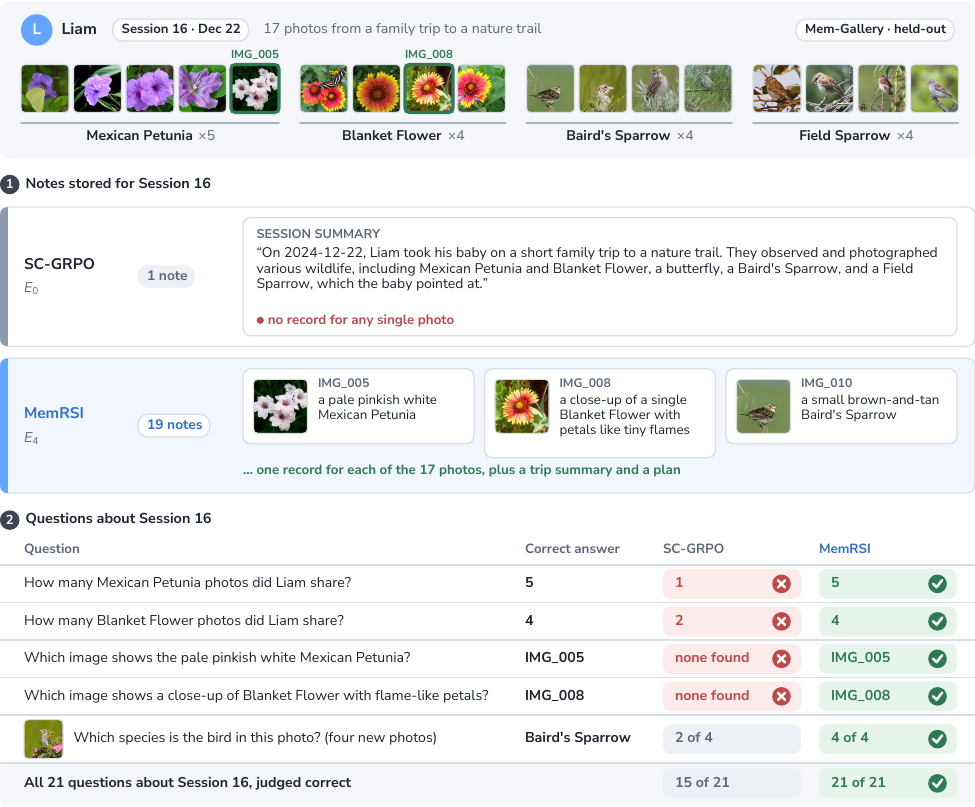}
  \caption{\textbf{Photo-level records on a held-out Mem-Gallery history.} Top: the 17 photos Liam shares in Session~16; green frames mark the two photos that questions ask for by description. Middle: the notes each \role{manager} stores for the session (1). Bottom: questions about the session (2). The species question is asked once for each of four photos of the bird that are not in the history, so the rows include all six questions that \initname{} misses; the last row counts the 21 questions about the session with a judge score of at least 0.5.}
  \label{fig:photo_records_case}
\end{figure}

This section examines what the evolved \role{manager} retains compared with SC-GRPO ($E_0$), the policy that recursive evolution starts from.
Fig.~\ref{fig:stagewise_multimodal_case} follows one question on an unseen MemEye history through the stored note, the retrieved evidence, and the answer, Fig.~\ref{fig:memory_quality_cases} compares what the two policies store for a Mem-Gallery question, and Fig.~\ref{fig:photo_records_case} follows every question about one photo-rich Mem-Gallery session.
In all three cases, the stages share the same environment, retrieval, and \role{answer agent}, only the \role{manager} differs, and \methodname{} is the policy trained on Mem-Gallery.

\paragraph{Evidence recovery on an unseen history.}
In this MemEye history, the user shares three similar glucose dashboards in Session~6 and a follow-up dashboard in Session~11.
The question asks which Session-6 screen has the same time in range as the follow-up, which requires linking two images five sessions apart (Fig.~\ref{fig:stagewise_multimodal_case}).
Both \envname{} and $E_0$ fail across all four option orders, whereas \methodname{} answers all four correctly.
The answer options name the screens, and the question with its options is the retrieval query, so a note that keeps a screen's name can be retrieved for it.
$E_0$ stores the matching screen as a generic daily dashboard without its name, and of the two matching screens retrieval returns only the follow-up.
\methodname{} keeps the screen name, so retrieval returns both screens and the \role{answer agent} can compare them.
Keeping the name pays off only five sessions later, which is the kind of delayed value that $E_2$ credits by measuring each write on the final bank (Fig.~\ref{fig:harness_revision_outcomes}a).
The policy was never trained on MemEye, and the behavior that preserves the name was learned on Mem-Gallery.

\paragraph{Program recovery.}
In Fig.~\ref{fig:memory_quality_cases}, Liam shares four plotting programs and the plots of the first three, and the question shows the missing logarithm plot and asks which program draws it.
Retrieval matches the question text and the plot against the stored notes and their images.
$E_0$ attaches images to the three plots but stores each program as a text-only fact without its image, so retrieval returns the three plots but none of the four programs, and the \role{answer agent} picks the exponential plot.
\methodname{} stores each program as a visual record with its code image, retrieval returns all four programs, and the \role{answer agent} identifies the logarithm program.

\paragraph{Photo-level records.}
Fig.~\ref{fig:photo_records_case} comes from another held-out history, whose user is also named Liam.
In Session~16, he shares 17 photos from a nature trail: five Mexican Petunias, four Blanket Flowers, and two lookalike sparrow species with four photos each.
$E_0$ condenses the session into a single summary that names the species but describes no individual photo, whereas \methodname{} writes one record for each photo, with its image and what the conversation says about it.
The summary is enough to name what Liam saw but not to answer questions about individual photos: $E_0$ undercounts both flowers, finds no image for either described photo, and identifies the sparrow in only two of four new photos.
Over all 21 questions about the session, $E_0$ answers 15 correctly and \methodname{} answers all 21.
In all three cases, the evolved \role{manager} stores more of each observation (its name, its image, or a record of its own), and the later questions depend on exactly this information.

\FloatBarrier
\section{Error Analysis}
\label{app:failure_modes}
\begin{figure}[H]
  \centering
  \includegraphics[width=\textwidth]{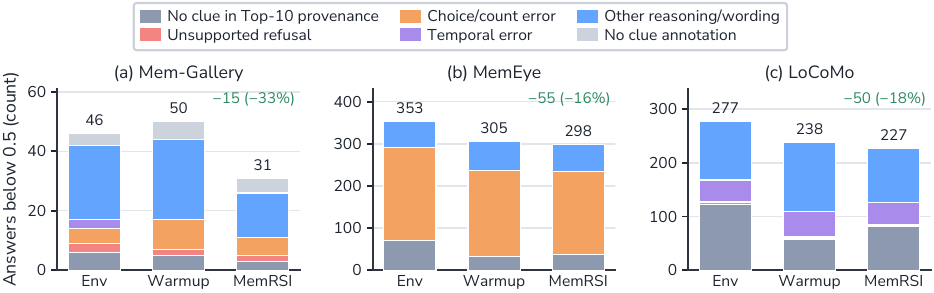}
  \caption{\textbf{Error types among answers with judge score below 0.5.} Each answer has one type. In the figure, ``No clue in Top-10 provenance'' denotes answers whose annotated source round was not retrieved, and ``No clue annotation'' denotes questions without an annotated source round. Labels give count changes from \envname{}. Memory comes from the Mem-Gallery-trained policy.}
  \label{fig:failure_analysis}
\end{figure}

After training, fewer answers fall below the judge threshold on all three benchmarks (Fig.~\ref{fig:failure_analysis}).
Among the remaining errors, visual choice and counting dominate on MemEye, while LoCoMo shows more temporal and answer-side reasoning errors.
Refusals on questions the evidence supports stay rare at every stage.
These error types arise when the \role{answer agent} must discriminate fine visual details or reason over time from the retrieved evidence, suggesting that part of the remaining gap lies on the answering side rather than in the stored memory.

\FloatBarrier
\definecolor{promptblue}{HTML}{63A4FF}
\definecolor{promptslate}{HTML}{8D99AE}
\tcbset{
  promptstyle/.style={
    enhanced, breakable,
    width=\linewidth,
    colback=promptblue!3, colframe=promptslate!55,
    boxrule=0.4pt, arc=2pt,
    left=7pt, right=7pt, top=5pt, bottom=5pt,
    colbacktitle=promptblue!13, coltitle=metafg,
    fonttitle=\sffamily\bfseries\small, toptitle=2pt, bottomtitle=2pt,
    before skip=8pt plus 2pt, after skip=10pt plus 2pt,
    before upper={\setlength{\parskip}{2.5pt}\setlength{\parindent}{0pt}%
      \raggedright\ttfamily\footnotesize},
  }
}
\newtcolorbox{promptbox}[1]{promptstyle, title={#1}}
\newcommand{\promptrole}[1]{\par\vspace{1pt}%
  \tcbox[on line, boxsep=0pt, left=3pt, right=3pt, top=1.5pt, bottom=1.5pt,
    colback=promptslate!20, colframe=promptslate!20, boxrule=0pt, arc=1.5pt]%
    {\normalfont\sffamily\scriptsize\color{metafg}#1}\par\nobreak}
\newcommand{\ph}[1]{\leavevmode{\color{promptblue!60!black}\{#1\}}}
\newcommand{\pskip}{\leavevmode{\color{promptslate}[\ldots]}}
\newcommand{\phead}[1]{\par\vspace{1pt}{\normalfont\sffamily\bfseries\footnotesize #1}\par\nobreak}

\section{Prompts}
\label{app:prompts}

Key excerpts of the prompts of the \role{manager}, revision \role{proposer}, \role{answer agent}, and judge are shown below; \pskip{} marks omitted passages, wording is condensed in places, and \ph{braces} mark inputs filled at run time.
In the \role{manager} prompt, entity\_overview, entity\_episode, and atomic memories form the State, History, and Records layers of the \bank{}.

\begin{promptbox}{Memory manager}
\promptrole{Instructions}
You are the memory manager of a multimodal long-term conversation memory system. \pskip{} Your task is to produce the memory operations justified by the new session.

\phead{Priorities, in order}
- Preserve useful information faithfully. Never destroy meaningful historical evidence.\\
- Store each independently useful coherent fact, event, list, decision, procedure, or state transition as a fine-grained atomic memory.\\
- Keep entity identity consistent across sessions.\\
- Represent meaningful corrections and state transitions explicitly.\\
- Preserve exact provenance and image metadata. \pskip{}

\phead{Three-layer memory design}
- Every substantive information unit in the incoming dialogue must be represented by at least one grounded atomic memory.\\
- For each main entity or coherent thread that meaningfully develops in the current session, ADD at most one entity\_episode memory describing what happened to or about that entity in this session.\\
- entity\_overview is a compact current-state anchor for a canonical entity. It is not the entity's event history and does not replace entity\_episode or atomic memories.

\phead{Operation rules}
ADD is the normal operation for new information. UPDATE is rare. \pskip{} Correction: the old claim was actually wrong. UPDATE the exact incorrect memory. Historical transition: both the old and new states were true at different times. Preserve the old historical memory. ADD a new state\_transition memory containing both states and the change time. \pskip{} Every useful image should normally produce one atomic visual\_observation memory. \pskip{}

\phead{Output}
\{"actions": [\{"operation": "ADD|UPDATE|DELETE|NOOP", "target\_memory\_indices": [], "memory": \{"text": \pskip, "metadata": \{"memory\_type", "main\_subject", "source\_dialogue\_ids", "source\_image\_ids", "temporal\_status", \pskip\}, "images": [\pskip]\}\}]\}

\promptrole{Input}
Existing memories: {[}\ph{index}{]} text: \ph{memory text}\quad metadata: \ph{type, subject, sources}\quad images: \ph{image fields}\quad \pskip\\
New incoming conversation: Observation \ph{index}\quad metadata: \ph{session, turn, timestamp}\quad text: \ph{turn text}\quad images: \ph{image fields}\quad \pskip
\end{promptbox}

\begin{promptbox}{Revision proposer}
\promptrole{User}
You are the sole scientific proposer for the next frontier candidate in a controlled Recursive Policy--Harness Evolution experiment for a multimodal personal memory manager trained with online GRPO.

Return exactly one high-confidence, bounded, experimentally testable change to the harness. This is not brainstorming, ablation enumeration, or a proposal portfolio. One proposal may couple several implementation details only when they realize one causal hypothesis. \pskip{} Use the appended reports and source as authoritative evidence. \pskip{} Choose the single approach best supported by the evidence.

Preserve \pskip{} the frozen evaluation. \pskip{} Specify the change mathematically or procedurally, \pskip{} and what formal result would retain it. Return one JSON object with the keys title, hypothesis, measured\_bottleneck, algorithm, \pskip

\ph{evidence packet: protocol, retained and rejected round reports, training configuration and metrics, harness source}
\end{promptbox}

\begin{promptbox}{Answer agent}
\promptrole{System}
You are an AI assistant evaluated on multimodal long-term conversational memory. \pskip{} If multiple pieces of information about the same event appear in the conversation, always rely on the most recent information. \pskip{} Only answer using information contained within the multimodal conversation. Do not hallucinate. Always remain consistent and grounded in the dialogue history.

\promptrole{User}
The retrieved memory contents are as follows: {[}Memory \ph{index}{]} timestamp: \ph{timestamp}\quad memory\_content: \ph{memory text}\quad image\_id: \ph{image id}\quad image\_content: \ph{image}\quad \pskip

Original source-round evidence for the retrieved memories follows. These are deduplicated original conversation rounds, ordered chronologically. Use them to verify the memory summaries; if a summary conflicts with its source round, prefer the original evidence. {[}Evidence Round \ph{dialogue id}{]} related\_memories: [\ph{memory indices}]\quad timestamp: \ph{timestamp}\quad \ph{original round}\quad \pskip

Your task is to answer the question \pskip{} with the help of memory content. \pskip{} The current question is as follows: \ph{question}
\end{promptbox}

\begin{promptbox}{Judge}
\promptrole{User}
You are an impartial judge evaluating the memory capabilities of an AI assistant with the question-answering task. Your task is to compare the Assistant's Answer against the Ground Truth and assign a score of 0, 0.25, 0.5, 0.75, or 1.

Score 0 (Incorrect / Miss): The answer contradicts the Ground Truth. \pskip\\
Score 0.25 (Poor / Tangential): The answer touches on the topic but misses the core entity or key value required. \pskip\\
Score 0.5 (Partial / Vague): The answer captures the main entity or concept correctly but misses a part of the required supporting details. \pskip\\
Score 0.75 (Good / Minor Imperfection): It misses only minor details that do not alter the main truth. \pskip\\
Score 1 (Correct / Exact): The answer is accurate, precise, and confident. \pskip

Question: \ph{question}\quad Ground Truth: \ph{reference answer}\quad Assistant Answer: \ph{model answer}\\
Output strictly in the following JSON format: \{"score": <0, 0.25, 0.5, 0.75, or 1>, "reasoning": "<short explanation>"\}
\end{promptbox}

\FloatBarrier
\end{document}